\documentclass[default,iicol]{sn-jnl}

\usepackage{moreverb,url}
\usepackage{subcaption}
\usepackage{amsmath}
\usepackage{graphicx}
\usepackage[linesnumbered,ruled,vlined,algo2e]{algorithm2e}

\jyear{2021}%

\theoremstyle{thmstyleone}%
\theoremstyle{thmstyletwo}%

\theoremstyle{thmstylethree}%

\begin{document}

\title[Article Title]{Design of an All-Purpose Terrace Farming Robot}



\author[1]{\fnm{Vibhakar} \sur{Mohta}}
\equalcont{These authors contributed equally to this work.}

\author[1]{\fnm{Adarsh} \sur{Patnaik}}
\equalcont{These authors contributed equally to this work.}

\author[1]{\fnm{Shivam Kumar} \sur{Panda}}
\equalcont{These authors contributed equally to this work.}

\author[1]{\fnm{Siva Vignesh} \sur{Krishnan}}
\equalcont{These authors contributed equally to this work.}

\author[1]{\fnm{Abhinav} \sur{Gupta}}
\equalcont{These authors contributed equally to this work.}

\author[2]{\fnm{Abhay} \sur{Shukla}}
\equalcont{These authors contributed equally to this work.}

\author[3]{\fnm{Gauri} \sur{Wadhwa}}
\equalcont{These authors contributed equally to this work.}

\author[1]{\fnm{Shrey} \sur{Verma}}
\equalcont{These authors contributed equally to this work.}


\author[1]{\fnm{Aditya} \sur{Bandopadhyay}}


\affil[1]{\orgdiv{Department of Mechanical Engineering}, \orgname{IIT Kharagpur}, \orgaddress{\state{WB}, \country{India}}}


\affil[2]{\orgdiv{Department of Industrial \& Systems Engineering}, \orgname{IIT Kharagpur}, \orgaddress{\state{WB}, \country{India}}}

\affil[3]{\orgdiv{Department of Aerospace Engineering}, \orgname{IIT Kharagpur}, \orgaddress{\state{WB}, \country{India}}}


\abstract{ Automation in farming processes is a growing field of research in both academia and industries. A considerable amount of work has been put into this field to develop systems robust enough for farming. Terrace farming, in particular, provides a varying set of challenges, including robust stair climbing methods and stable navigation in unstructured terrains. We propose the design of a novel autonomous terrace farming robot,  ‘Aarohi’, that can effectively climb steep terraces of considerable heights and execute several farming operations. The design optimisation strategy for the overall mechanical structure is elucidated. Further, the embedded and software architecture along with fail-safe strategies are presented for a working prototype. Algorithms for autonomous traversal over the terrace steps using the scissor lift mechanism and performing various farming operations have also been discussed. The adaptability of the design to specific operational requirements and modular farm tools allow Aarohi to be customised for a wide variety of use cases.}





\maketitle

\section{Introduction}
According to the economic survey 2020-21 \cite{GDP}, the share of agriculture in the gross domestic product (GDP) is only about 20.19\% even though agriculture is the primary source of livelihood for 58\% of India’s population. This can highly be regarded to the labour-intensive practices being followed in Indian agriculture due to the lack of resources. On steep slopes, terrace farming is one of the most efficient ways of conserving soil and water \cite{engdawork2014long}. It helps in reducing runoff by over 41.9\% and sediment loss by over 52\% \cite{deng2021advantages}. By 2050, it is projected that the food demand of India will almost double \cite{hamshere2014india}. To meet such demands, the need for revolutionising the field of agriculture arises, which can be achieved by automating certain agricultural practices. The agriculture sector has a 57\% potential for automation and was ranked 4th in a McKinsey report that assessed the automation potential of major global sectors \cite{mckinsey}. Thereby this sector presents a great opportunity for reducing the cost of production and increasing productivity. However, there are multiple challenges in automating the process of terrace farming. The use of petrol or diesel tractors is non-viable due to the narrow size of steps \cite{spugnoli2013environmental}. As a result, most of the farmers rely on the use of traditional hand tools by abundant household labor\cite{tiwari2004implications} \cite{varisco1991future}. Moreover, socio-economic challenges, such as poverty and illiteracy, further exacerbate the problem of automating terrace farming \cite{chapagain2017agronomic}.

Over the last few decades, significant advances in sensor technology and computer capacity have paved the path for autonomous systems to be deployed in a variety of sectors \cite{10.1145/3387304.3387321, 9265493, 9476722, singhal2019real}. Likewise, the challenges faced in the agricultural sphere have been addressed in the literature using drones, robotic arms, and autonomous vehicles. Ladybird \cite{Ladybird}, an autonomous field robot was designed at the Australian Centre for Field Robotics to collect data in an agricultural field. An autonomous mobile plant irrigation robot, using wireless communication to communicate with a moisture sensor, was developed and tested \cite{xiong2020autonomous}. Conventional methods of seed sowing, using animals or tractors, are time and energy-consuming. To tackle this problem at a low cost, a swarm of robots with minimal individual intelligence was developed for sowing tasks \cite{blender2016managing}. A centralised coordinator, OptiVisor is used here for path planning and optimisation. Another autonomous robot prototype developed particularly for the task of seed sowing was developed in \cite{naik2016precision}. Autonomous robots for sweet pepper harvesting in greenhouses \cite{arad2020development}, apple harvesters\cite{silwal2017design} have been developed using fruit detection algorithms, path planning, localisation and motion control. Conventionally, pesticide spraying is done by the means of backpack sprays or tractor sprayers. A human-robot collaborative method of spraying specific targets \cite{berenstein2017human} was developed using different image based targeting algorithms. These studies have been done to automate a particular process in the process of agricultural production, but there is a need to integrate such automated processes into one. Numerous studies have developed multipurpose agricultural robots \cite{sowjanya2017multipurpose, gollakota2011agribot} which can perform all agricultural tasks but are capable of traversing plain fields only. To overcome the challenge of uneven terrains and steps, drones have been used in terraces for pesticide spraying and crop health monitoring \cite{kong2018autonomous}. Studies to understand the application of machine learning, image processing and pattern recognition in drones for agricultural purposes have also been  conducted\cite{kulbacki2018survey}. However, there are multiple challenges posed by the usage of drones in agriculture such as high cost, and the inability to perform tasks like ploughing and harvesting. 

Full-scale development and commercialisation of a terrace farming robot is inherently a challenging task. Firstly, it requires the robot to be able to traverse long and curved individual terraces on a hilly slope safely and efficiently. The terraces in India are typically 60-80m long with varying widths along the length. This necessitates the farm robot to be as compact as possible to safely pass through these converging terraces. Active mapping of the environment is a primary requirement for all autonomous field robots, more so in the case of a terrace farming robot where a minor misjudgement in step dimensions can cause the robot to fall from huge heights. Secondly, for the commercial success of the robot, both the initial cost and the running cost of the system should be kept as low as possible keeping in mind the low income of farming households \cite{farmer_income}. The third and the most important consideration while designing such a terrace farming robot is to develop an effective mechanism to safely climb up the terraces whose step heights can easily reach up to 40 cm.

To this end, we reviewed several stair-climbing mechanisms proposed by researchers and engineers to date. Broadly, various stair-climbing robots can be put into either of these categories: (a) Wheeled robots, (b) Tracked or rail-based robots and (c) Legged robots.

Wheeled stair-climbing robots typically use multiple wheels in combination to roll, crawl and finally climb the stairs \cite{Siegwart1998DesignAI}. MSRox proposed by Mohsen \textit{et al.} is one such wheeled robot \cite{Dalvand2006StairCS}. The authors introduced a unique star-wheel design which consisted of an assembly of three wheels powered by electric motors on a freely rotating base. This configuration with a star-wheel showed good adaptability on rugged terrain and stairs, however, it suffers from a major flaw. The radius of the star wheel should be significantly greater than the step height for the robot to climb the stairs. This constraint makes the entire system bulky and complex. While it may work well for small stair heights (10-17 cm), it can not be used on terrace farms that have small terrace widths and steep heights. Moreover, the chassis of the robot is also very unstable for supporting huge weights required for terrace farming.

Tracked robots are also a popular choice for stair-climbing as they have been found to work well for rough terrains \cite{Lee2009ASR, Michaud2003AZIMUTAL, Kim2010WheelH}. Suyang \textit{et al.} designed a wheelchair robot with variable geometry tracked mechanism for assistance in stair climbing \cite{Yu2010ConfigurationAT}. The robot could adapt to convex terrain by active control of tension in the track. Yoon-Gu \textit{et al.} also developed a terrain adaptive wheel track hybrid robot that used both wheel and track-based systems \cite{Kim2012AutonomousTA}. However, similar to wheeled robots, both of them suffer from large size and complexity which makes them impractical to use in narrow, steep terraces.

Humanoids, such as HONDA ASIMO \cite{1389593}, NAO \cite{6094533}, and HRP-2 \cite{4399104}, have demonstrated capability for stair climbing but they are extremely expensive and difficult to put into farming use. On the other hand, Legged robots such as RHex \cite{Saranl2001RHexAS} could be augmented with farming tools easily but the unstable chassis motion discourages their use for farming applications that require high precision and accuracy.
\begin{figure*}[!tbh]
    \centering
    \includegraphics[width=0.75\linewidth]{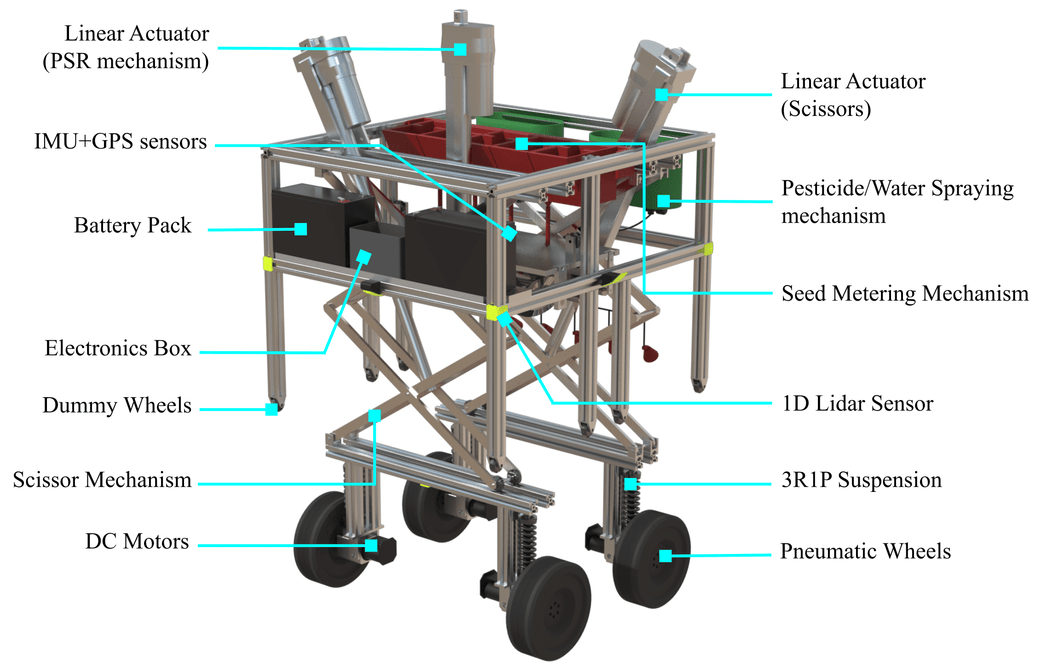}
    \caption{Aarohi’s Robot Model}
    \label{fig_robot_model}
\end{figure*}
\begin{figure*}
    \centering
    \begin{subfigure}[b]{0.49\textwidth}
         \centering
         \includegraphics[width=\textwidth]{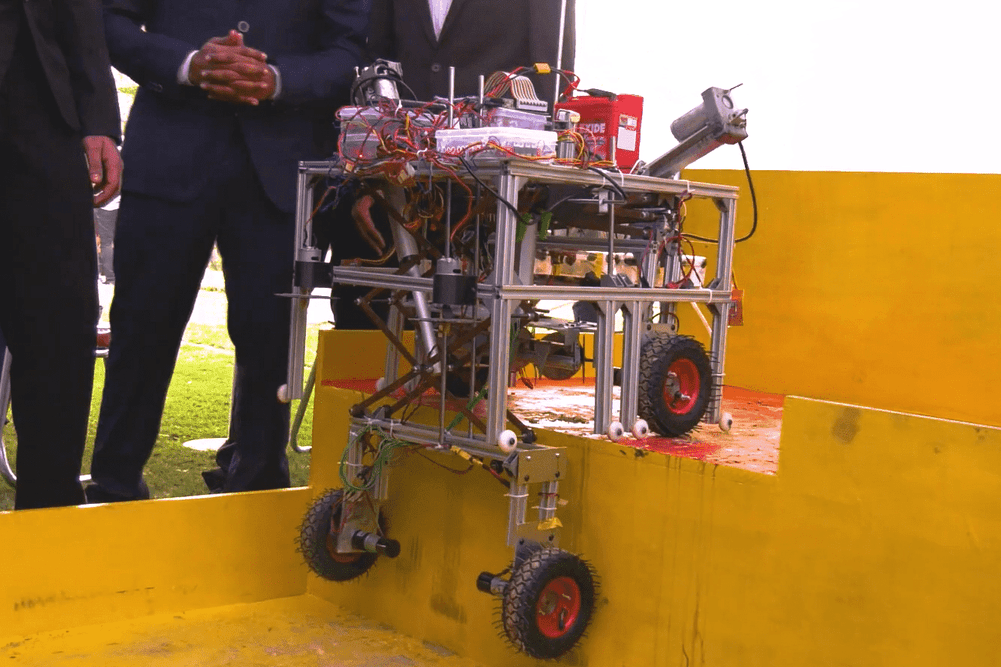}
         \caption{Prototype with extended scissors}
         \label{fig_robot_pic1}
     \end{subfigure}
     \begin{subfigure}[b]{0.49\textwidth}
         \centering
         \includegraphics[width=\textwidth]{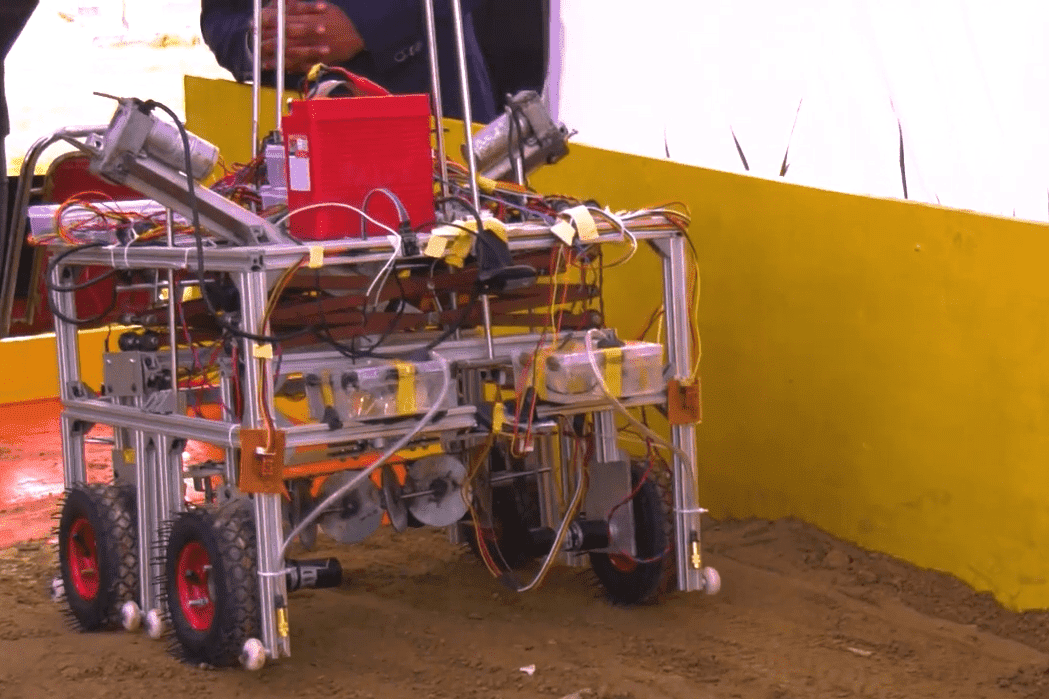}
         \caption{Prototype with closed scissors}
         \label{fig_robot_pic2}
     \end{subfigure}
         \caption{Images of the robot prototype in a test arena}
    \label{fig:real_robot}
\end{figure*}

Apart from the aforementioned categories, there are other robots proposed by researchers which can be used for stair climbing. Zhang \textit{et al.} proposed the design for a stair-cleaning robot that combined lift or drop motion with the telescopic linear motion to achieve stair climbing \cite{Zhang2016ANC}. However, the long telescopic mechanism accounts for large weight and size which not only restricted its quick turning capability but also imbalanced the entire robot in one of the stair descent configurations. Later, the design was modified to a small leg-wheel hybrid robot which showed promising results for cleaning applications on small stairs \cite{Zhang2016ANC}. Unfortunately, the entire system’s load is supported by a single actuated joint in one of the configurations, which would make the robot highly unstable if this mechanism were replicated for terrace farming with heavy payloads.

From our discussion so far, it can be ascertained that there is a lack of agriculture robots that can provide complete automation of common farming processes such as ploughing, sowing, watering, etc. Moreover, existing mechanical designs of stair climbing robots can not be directly put to use for building a terrace-climbing farming robot. Therefore, to overcome this issue, we propose the novel design of ‘Aarohi’, an autonomous terrace farming robot. The key innovations associated with this work are summarised as follows: 
\begin{enumerate}
    \item Novel scissor lift based mechanism for stair climbing to handle steep terrace steps relevant to terrace farming scenarios
    \item Re-configurable design for attaching and detaching different modules for ploughing, sowing, pesticide spraying, watering etc.
    \item Autonomous solution to traverse the terrace and perform a multitude of agricultural activities.
\end{enumerate}

\begin{figure*}[!tb]
    \centering
    \begin{subfigure}[b]{0.49\textwidth}
         \centering
         \includegraphics[height=0.8\textwidth]{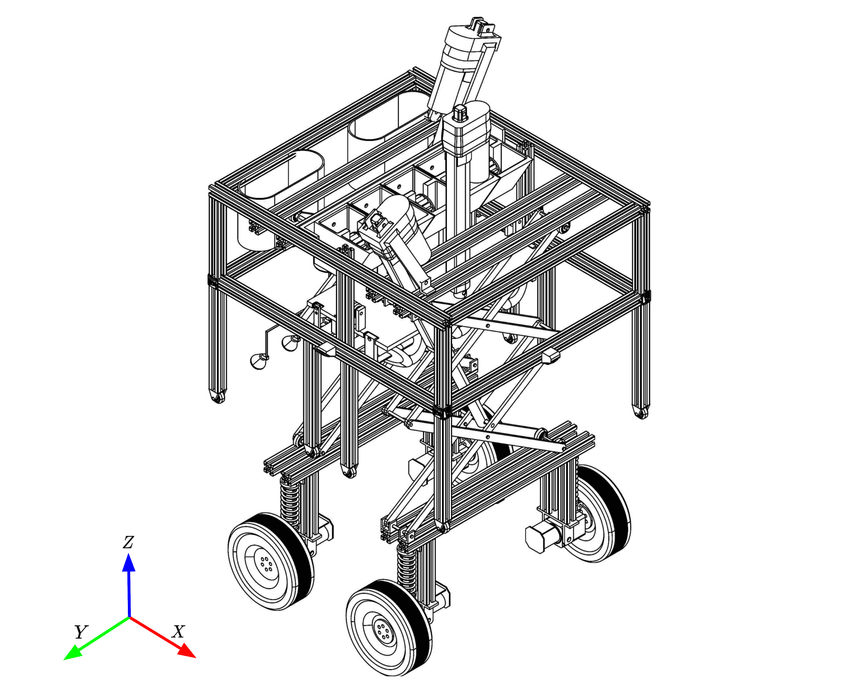}
         \caption{Design with two pairs of scissors}
         \label{fig_two_pair_wheels}
     \end{subfigure}
     \begin{subfigure}[b]{0.49\textwidth}
         \centering
         \includegraphics[height=0.8\textwidth]{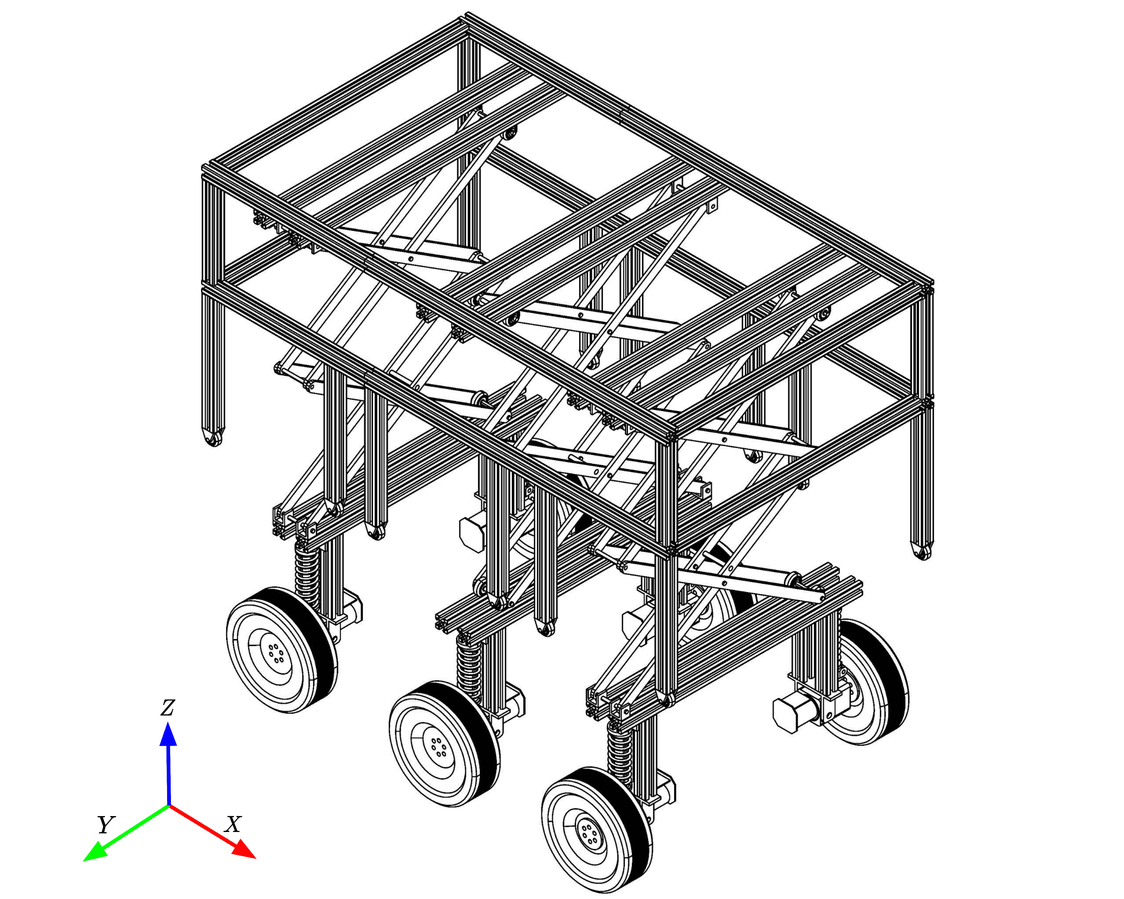}
         \caption{Design with three pairs of scissors}
         \label{fig_three_pair_wheels}
     \end{subfigure}
     \caption{Design with varying wheel pairs}
    \label{fig:wheel_pairs}
\end{figure*}

The following aspects are needed to be emphasised before proceeding further:

\begin{enumerate}
    \item Only the generalised design, design process, control and software architecture has been proposed in this work. The specifications of hardware mentioned in this paper are for the prototype developed by us and are in no way exclusive to the design.
    \item Steps and stairs mentioned are the steps in the terrace farming region and are interchangeably used throughout the paper. It is also assumed that the steps are flat and not ramped.
    \item It is assumed that the robot is not operated in a heavy windy environment.
    \item The number of pairs of powered wheels and the number of scissor levels used for climbing is prototype specific. It could be extended to $n$ number of wheels and $m$ scissor levels depending on the size of the robot, where $n \geq 2$ and $m \geq 2$. Fig. \ref{fig_two_pair_wheels} and fig. \ref{fig_three_pair_wheels} shows the design of the chassis and stair climbing module with two and three pairs of wheels, respectively.
\end{enumerate}

The rest of the paper is organised as follows. Section \ref{Mechanical_Design} introduces the overall design and discusses the mechanical structure of the entire assembly and the design optimisation process in detail. Finally,  section \ref{Software_Architecture} describes the software architecture of the robot in-depth, including navigation strategies, stair climbing strategy and control of various modules. All the results are finally concluded in section \ref{Conclusions}.

\section{Mechanical Design}
\label{Mechanical_Design}
The novel modular design enables the robot to traverse the terrain of terrace farming steps, climb up and down the terrace steps, and perform various farming operations such as ploughing, sowing, watering, pesticide spraying, etc. The scissor lift mechanism allows the robot to climb up and down the steps. Independent detachable modules have been included in the robot to perform basic farming activities. The advantage of having separate modules is that these can be attached or detached to the robot depending upon the work at hand. Also, since each module works independently of the others, it makes it easier to repair or replace any module in case of damage. 
\par During prototyping, V-slotted aluminium extrusions were used to build the chassis of the robot. The locomotion is carried out by a four-wheel differential drive. This drive mechanism allows the robot to turn about its position, change its orientation during a task and proceed in the desired direction. Pneumatic wheels are employed for the drive as they provide higher load capacity and better traction while traversing on different kinds of soil. To prevent any damage to the mechanisms inside the prototype due to vibrations, an independent 3-revolute-1-prismatic (3R1P) suspension mechanism with a coaxial spring cum damper system has been attached to each wheel. Dummy wheels that are not powered are attached to the chassis in order to help the robot climb the steps.


\begin{table*}[!tb]
    \centering
    \caption{Parameters in scissor-lift mechanism design}

    \begin{tabular}{c p{5.3in} }
    \toprule
    \textbf{Parameter} & \hspace{2in}\textbf{Description}\\
    \midrule
    $n$ & Number of scissor levels\\
    $\theta$ & Angle made by a positive sloping link with x-axis\\
    $D$ & Length of the scissor-link\\
    $h$ & Height of the scissor lift mechanism in its current configuration\\
    $H$ & Height of the stairs that the robot needs to climb\\
    $l$ & Length of the actuator AB\\
    $S$ & Stroke length of the linear actuator\\
    $i$ & Number of scissors below the point B\\
    $a$ & Fractional number between $(0,0.5]$. $aD$ denotes the length PA\\
    $b$ & Fractional number between $(0, 0.5]$. $bD$ denotes the length OB\\
    $F$ & Linear actuator force required to actuate the scissor-lift at current configuration\\
    $L$ & Load lifted by the scissor-lift, incorporating both robot and scissor-lift weight\\
    $t$ & Thickness of the scissor-link\\
    \bottomrule
    \end{tabular}
    \label{tab:Var_scissor_lift}
\end{table*}

\subsection{Stair Climbing Mechanism}
The scissor lift mechanism allows the robot to climb up and down the step. It also facilitates variable height adjustment during watering, pesticide spraying, and harvesting. Only two pairs of scissors are used in the prototype developed. However, multiple pairs of scissors could be used for climbing depending upon the dimensions and weight of the robot. For example, Fig. \ref{fig_three_pair_wheels} shows a modified design with three pairs of scissors.
\par The pair of front wheels and rear wheels are attached to one scissor lift mechanism each. These two scissor lift mechanisms are differentially actuated to climb up and down the stairs. There are four pairs of dummy wheels -- one ahead of the front wheel, two between the front and rear wheels and one behind the rear wheels -- to support the climbing motion. The number of dummy wheels can be changed according to the dimension of the robot and the number of scissor lift pairs used in the robot. Scissors connecting to the front and rear pair of wheels are attached with one linear actuator each to support the lifting of the robot. Guide rails are attached to the scissors to ensure proper vertical lift motion. The stair climbing and descending algorithms are discussed in detail in section \ref{subsec_stair_climb}.
\par The scissor-lift mechanism, which is considered as one of the most stable mechanisms for lifting actions in industries, has been used instead of fork-lift or lead-screw for lifting the robot because of the relative easiness in manufacturing and assembling.

\subsubsection{Design of scissor-lift mechanism}

When designing a scissor-lift mechanism, the location of the linear actuator’s attachment is critical because it directly impacts the potency of the system, that is, the mechanical advantage and the velocity ratio, and thereby influencing the choice of the linear actuator to actuate the system, and hence, the overall cost. In short, the position variables of the linear actuator attachment affect the force output and the stroke length of the actuator. There are other design parameters such as the link lengths, link thickness, number of scissors, etc. that directly influence the force output and the stroke length of the actuator.

\begin{figure}[!tb]
    \centering
    \includegraphics[width=0.95\linewidth]{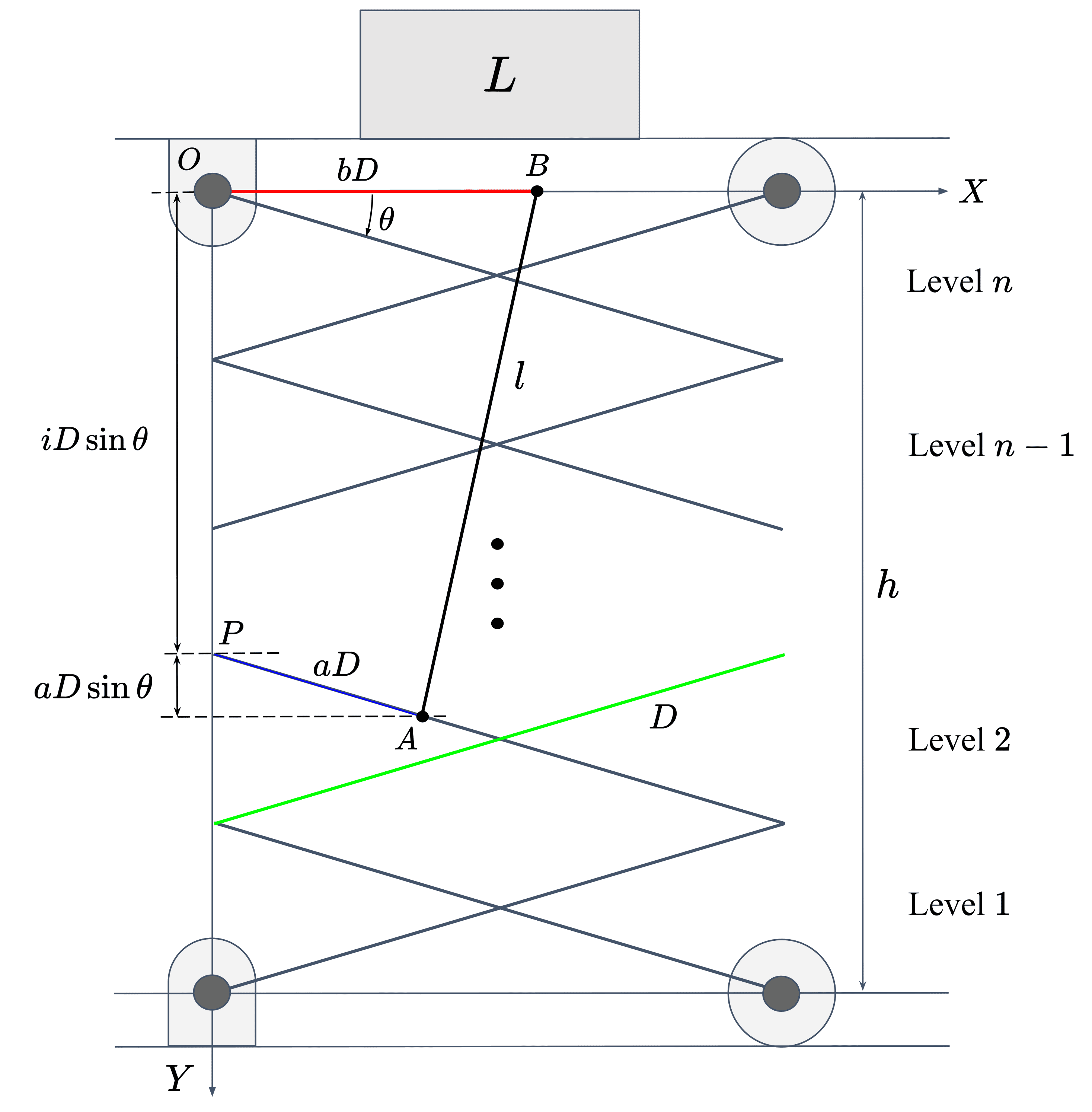}
    \caption{Representation of an n-level scissor lift mechanism with the variables used.}
    \label{fig_scissor_lift}
\end{figure}

Before picking an ideal set of position variables, it is necessary to establish equations for force output and the stroke length of a linear actuator in terms of the scissor-lift position variables. Without loss of generalisation, we can consider a $n$ stage scissor lift mechanism as shown in Fig. \ref{fig_scissor_lift}. The linear actuator has one end static and the other translating. The translating end is attached to a point on a positively sloping link in the scissors and the static end is attached to a fixed support. When the translating end of the actuator extends, it will cause the scissor-lift to extend and vice versa.

As shown in Fig. \ref{fig_scissor_lift}, A is the point where the translating end of the linear actuator is attached to the scissor and hence, it is the point where force is applied. B is the point where the static end of the linear actuator is attached. O is the origin and XY is the reference co-ordinate axes for the analysis. Various symbols used and their corresponding descriptions are tabulated in Table \ref{tab:Var_scissor_lift}.

\begin{figure*}[!tb]
    \centering
    \begin{subfigure}[b]{0.49\textwidth}
        \centering
        \includegraphics[width=\linewidth]{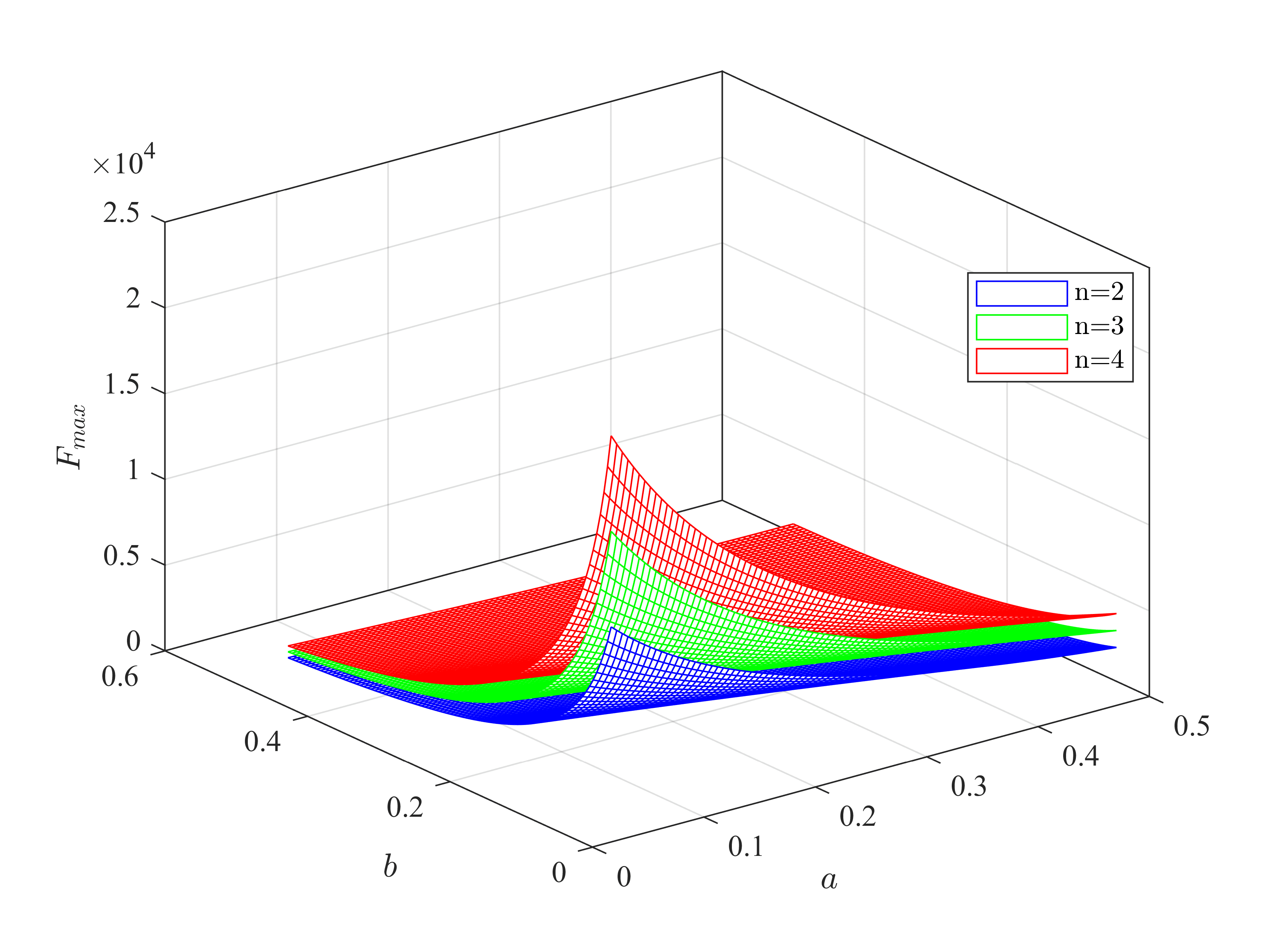}
        \caption{Variation of maximum Force output with position variables $a$, $b$ and number of scissors $n$}
        \label{fig_F_vs_a_b_n}
     \end{subfigure}
     \begin{subfigure}[b]{0.49\textwidth}
        \centering
        \includegraphics[width=\linewidth]{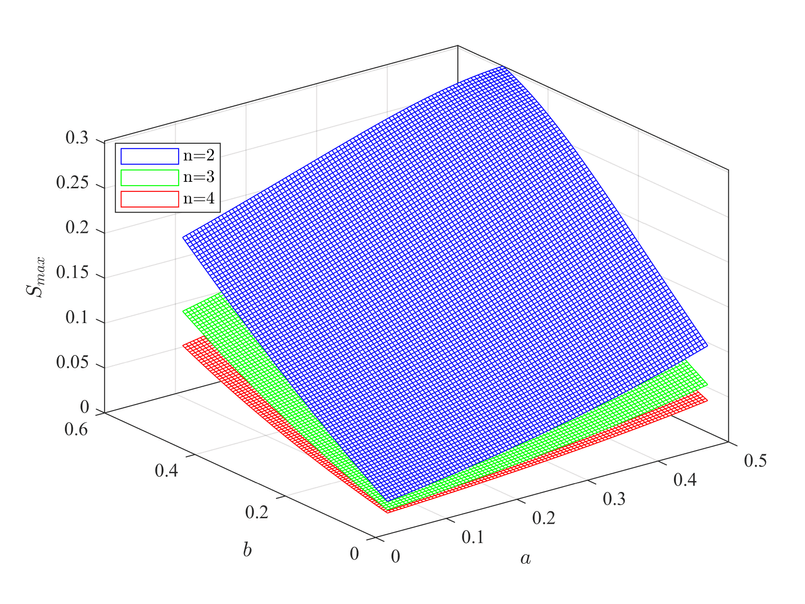}
        \caption{Variation of the maximum stroke length with position variables $a$, $b$ and number of scissors $n$}
        \label{fig_S_vs_a_b_n}
     \end{subfigure}
\end{figure*}

Here, $a$, $b$, $\theta$ and $n$ are the only independent variables. $D$ and $i$ are assumed to be constant in our analysis. $i$ is set equal to $n-1$. The expressions $h$, $l$ and $F$, derived in \cite{scissor-lift}, in terms of the independent variables, are shown in (\ref{h}), (\ref{l}), and (\ref{f}), respectively. The stroke length of the linear actuator can be obtained by subtracting the length of the actuator in the maximum extended state and the maximum compressed state as expressed in (\ref{S}). It is also required to find out the maximum force required to actuate the scissor-lift to choose the linear actuator. The force required to actuate the scissor-lift will be maximum when the scissors are in the most compressed state, i.e., when $\theta =  \theta_{min}$. The maximum force, $F_{max}$, can be expressed as in (\ref{F_max}). The variation of $F_{max}$ and $S$ with respect to $a$, $b$, and $n$ is clearly illustrated in Fig. \ref{fig_F_vs_a_b_n} and Fig. \ref{fig_S_vs_a_b_n}, respectively. With an increase in both $a$ and $b$, $F_{max}$ decreases and $S$ increases. Also, with a increase in $n$, $F_{max}$ increases and $S$ decreases.
\begin{gather}
    h = n D \sin \theta \label{h} \\
    l = D \left[\lambda\cos^2{\theta} - 2b\overline{a}\cos{\theta} + b^2 + (i+a)^2 \right] ^{\frac{1}{2}} \label{l}\\
    F=nL\left[\frac{\lambda \cos^2{\theta}- 2b\overline{a}\cos{\theta} + b^2 + (i+a)^2 }{ \left( b\overline{a}\tan \theta - \lambda \sin \theta \right)^2} \right]^\frac{1}{2} \label{f}\\
    F_{max} = F\vert_{\theta =  \theta_{min}} \label{F_max} \\
    S = l\vert_{\theta = \theta_{max}} - l\vert_{\theta =  \theta_{min}} \label{S}
\end{gather}
Where $\lambda=\left(\overline{a}^2 - (i+a)^2\right)$ and $\overline{a}=1-a$. Clearly, $F_{max}$ and $S$ are conflicting functions with respect to $a$ and $b$. Hence, we formulate this problem as a multi-objective optimization problem, where the conflicting objectives, $F_{max}$ and $S$, are to minimized. The problem is mathematically formulated as shown in (\ref{multiobj}). 
\begin{figure}[!tb]
    \centering
    \includegraphics[width=\linewidth]{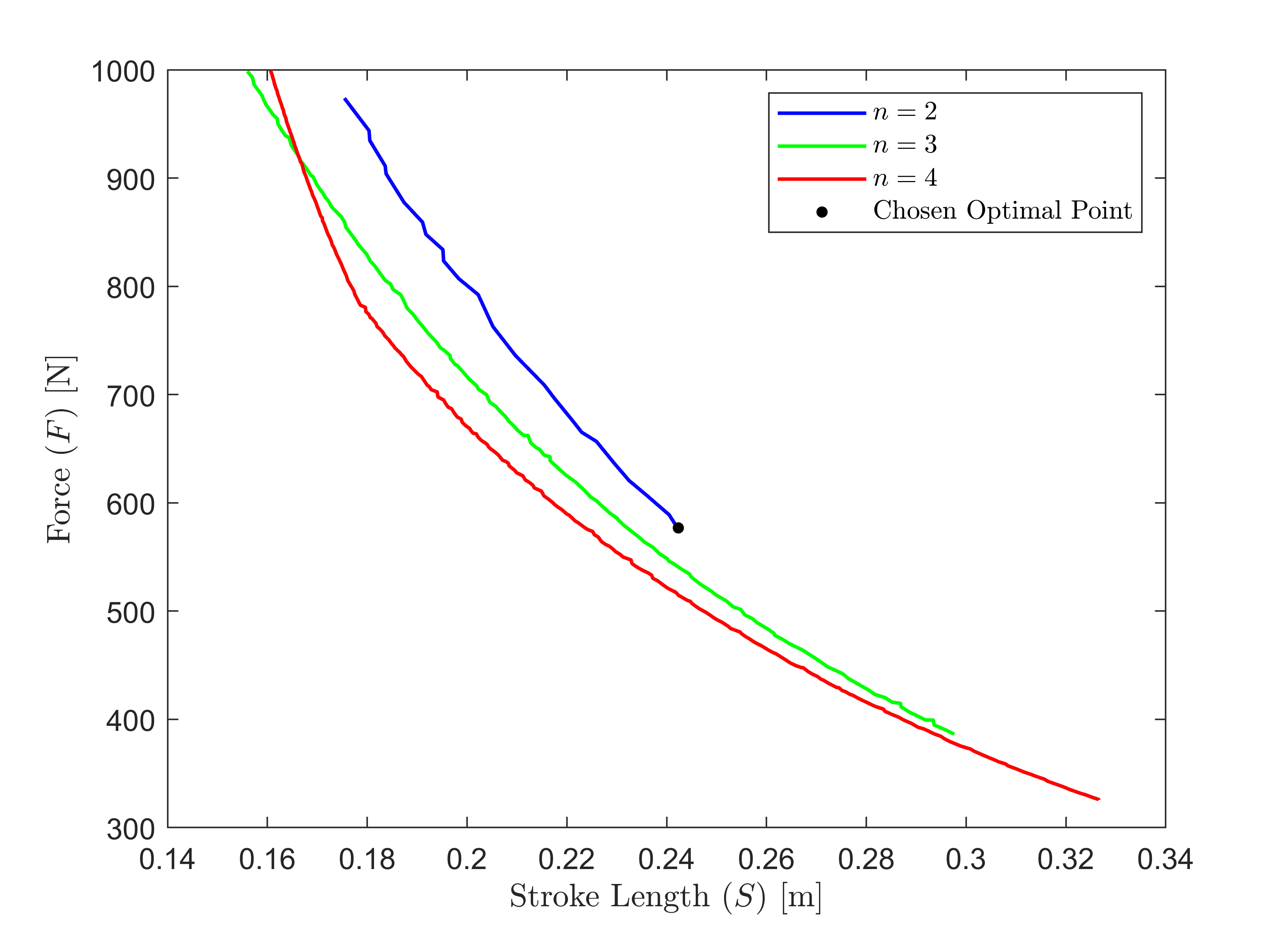}
    \caption{Pareto Front showing various values of $F_{max}$ and $S$ for different values of $n$}
    \label{fig_pareto_front}
\end{figure}
\begin{equation}
\begin{aligned}
\min_{a,\;b,\; \theta_{min},\; n,\; D,\; i} \quad & f = (F_{max}, S)^T\\
\textrm{S.T} \quad & 0 < a,b \leq 0.5\\
  & 5^\circ \leq  \theta_{min} \leq 10^\circ \\
  & D_{min} \leq D \leq H \\
  & i = n-1
  \label{multiobj}
\end{aligned}
\end{equation}
Here, $D_{min}$ is the minimum link length required to climb the step of height, $H$, calculated as shown in (\ref{D_min}).
\begin{equation}
     D_{min} = \frac{H}{n(1-\sin ( min(\theta_{min}) ))}
\label{D_min}
\end{equation}
Since a single solution cannot optimise all the conflicting objectives simultaneously, we seek to optimise the objective functions and present the Pareto Front. The Decision Maker, then, can choose a linear actuator from the Pareto Front solution set based on the use case of the robot. 

Multi-objective evolutionary algorithms (MOEAs) have grown in popularity in recent years as a result of their ability to find many Pareto-optimal solutions in a single simulation run. Algorithms such as Non-Dominated Sorting Genetic Algorithms (NSGA-II) \cite{996017}, Multi-Objective Grey Wolf Optimizer (MOGWO) \cite{mirjalili2016multi}, Multi-Objective Particle Swarm Optimizer (MOPSO) \cite{reyes2006multi}, Multi-Objective Bonobo Optimizer (MOBO) \cite{das2020multi} are widely used to obtain Pareto-front in real-world multi-objective problems.

\begin{table}[h]
    \centering
    \caption{Linear actuator specifications and position variables for the developed prototype.}
    \label{tab:lin_act_specs}
    \begin{tabular}{c c c}
    \toprule
    {Parameter} & {Datasheet Values} & {Final values used}\\
    \midrule
    $F_{max}$ & 576.840 N & 1000 N \\
    S & 0.242 m & 0.25 m\\
    a & 0.499 m & 0.5 m\\
    b & 0.158 m  & 0.16 m\\
    $\theta_{min}$ & $10^\circ$ & $10^\circ$\\
    D & 0.409 m & 0.4 m\\
    n & 2 & 2\\
    t & N/A & 3 mm Steel \\
    \bottomrule
    \end{tabular}
\end{table}
Here, we are employing \textit{gamultiobj}, a MATLAB package that uses a controlled, elitist genetic algorithm, to solve the constrained multi-objective optimisation problem (\ref{multiobj}). We set the population size to be $200$ and the Pareto fraction as $0.7$. We designed a robot for climbing stairs of height up to $40$ cm. In our case, choosing $n>4$ seemed sub-optimal and hence, we added an additional constraint on the number of scissors, $n = \{2, 3, 4\}$. The load $L$ moved by the scissor-lift and the height $H$ of the stair is fixed as 250 N and 40 cm, respectively. Solving the optimisation problem separately for each value of $n$, three different Pareto optimal fronts were obtained as shown in Fig. \ref{fig_pareto_front}.
\begin{figure}[!tb]
    \centering
    \includegraphics[width=\linewidth]{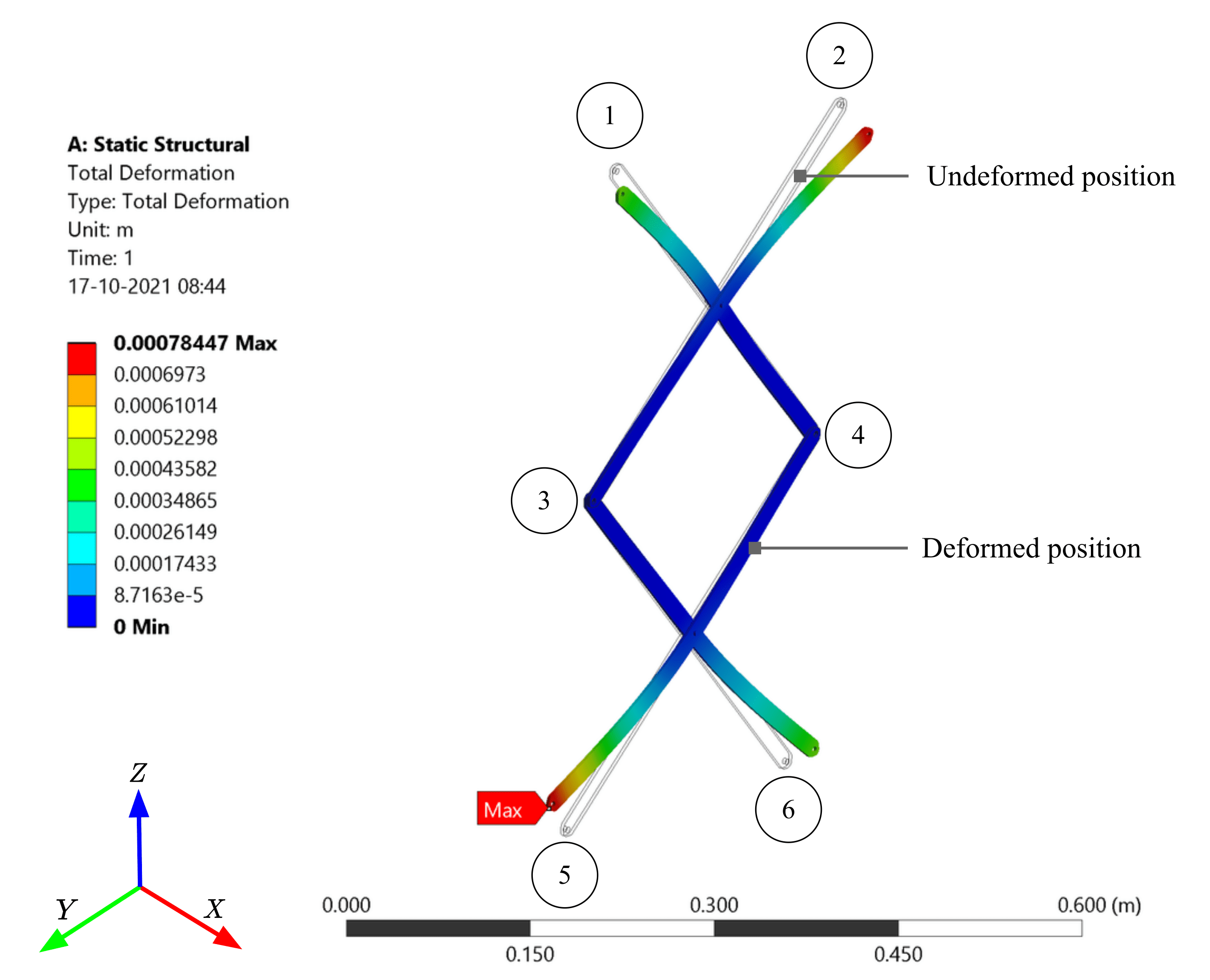}
    \caption{Simplified Model of the scissor-lift mechanism both in the undeformed and deformed position. Forces and torque are applied at ends marked as 1, 2, 5, and 6. Joints marked as 3 and 4 are fixed during the simulation.}
    \label{fig_ansys_model}
\end{figure}
\begin{figure}[!tb]
    \centering
    \includegraphics[width=\linewidth]{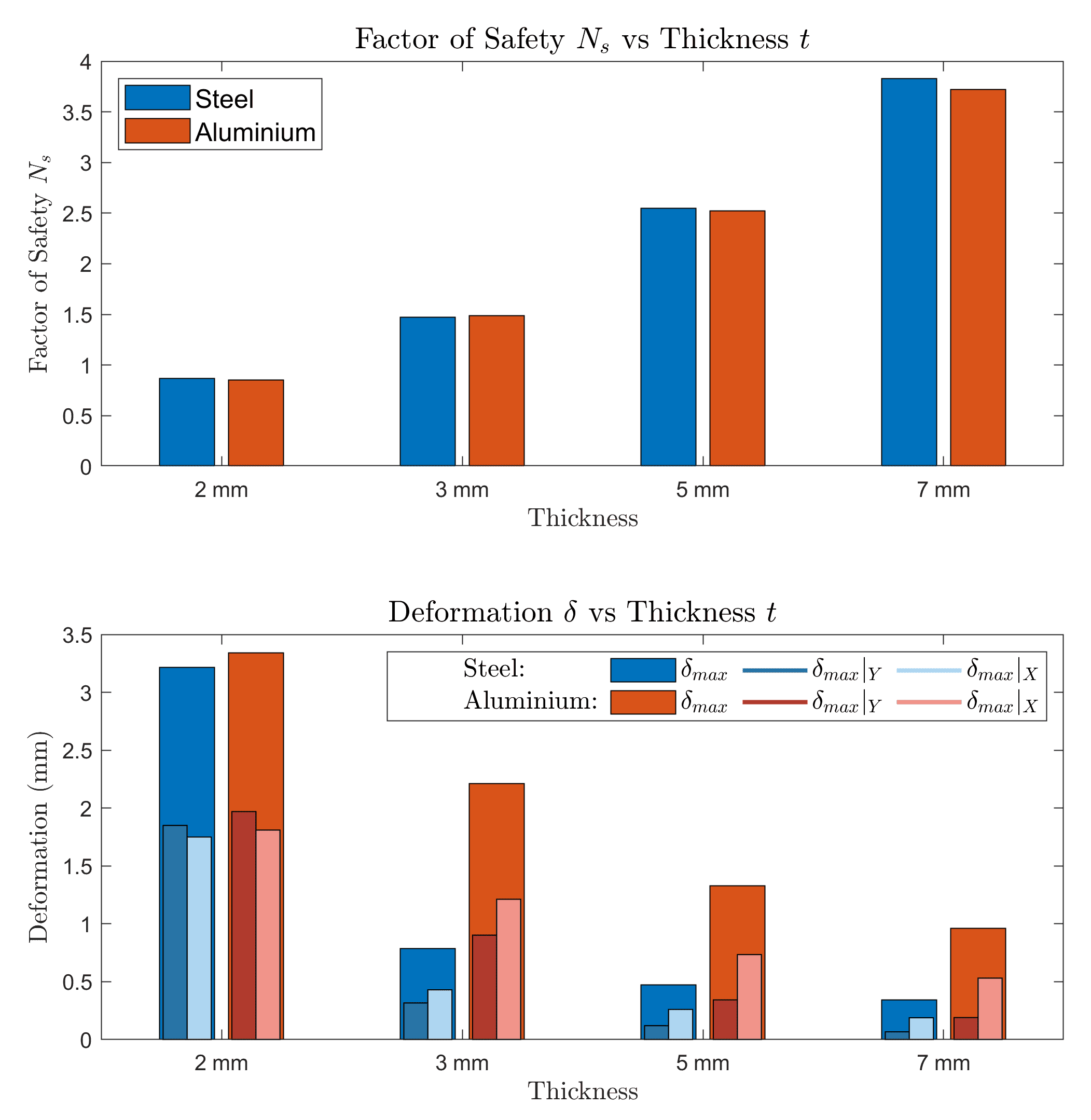}
    \caption{(a) Plot showing the variation of the factor of safety against various thicknesses considered for steel and aluminium. It can be observed that the values are similar for both the metals at a particular thickness; (b) Plot showing the variation of maximum deformation ($\delta_{max}$) and its components along X- ($\delta_{max}\vert_X$) and Y- ($\delta_{max}\vert_Y$) direction against various thickness considered for steel and aluminium.}
    \label{fig_ansys_plot}
\end{figure}
We have added additional constraints on the maximum force ($F_{max} \leq 1000$ N) and stroke length ($S \leq H$) to filter out the non-feasible values. The filtered values of $F_{max}$ and $S$ along with the corresponding independent variables have been tabulated:  \href{https://drive.google.com/file/d/14_180iAj2aL5CX9jh-V_5XnIY1kC9u17/view?us=sharing}{(LINK)}. The Decision Maker can use the table as a datasheet to compare them to other linear actuators on the market and select one based on price and availability. In our case, we chose a linear actuator with $25$ cm stroke length and $1000$ N maximum load. This choice of linear actuator introduced a factor of safety of approximately $1.74$ in the design. The final scissor-lift design parameters along with the linear actuator specifications chosen for the robot prototype are tabulated in Table \ref{tab:lin_act_specs}.

Further, the thickness of the scissor links needs to be optimised between maximum stress, deformation due to bending and buckling. Also, while optimising the thickness, the space for the placement of the linear actuators between the scissors has to be maximised. Given the problem, a simplified model of the scissor lift mechanism with two crosses used in the prototype was imported to ANSYS 17.1 for structural simulation analysis as shown in fig. \ref{fig_ansys_model}. This model refers to the maximum extended position of the scissor lift mechanism. In the simulation, considering 40 kg as the weight of the robot, respective forces and torques were applied to the ANSYS model at the ends marked as 1,2,5 and 6 in fig. \ref{fig_ansys_model}. The joints indicated as 3 and 4 in fig. \ref{fig_ansys_model} are fixed during the simulation. Two commonly used materials -- Steel and Aluminium, each with four commercially available and relevant thicknesses -- 2 mm, 3 mm, 5 mm and 7 mm -- were selected for the simulation study.

The factor of safety due to the maximum stress ($N_S$), and the absolute maximum total deformation ($\delta_{max}$) and its components in X - ($\delta_{max}\vert_X$) and Y-direction ($\delta_{max}\vert_Y$) were analysed during the ANSYS simulation study. Figure \ref{fig_ansys_plot} shows the values of parameters analysed during the study. It is to be noted that $\delta_{max}\vert_X$ physically signifies the deformation due to bending and buckling of the scissor links. 
It was found that the safety factors ($N_S$) for all the four thicknesses considered were similar for both metals. However, aluminium showed a higher maximum deformation than steel with the same thickness both in X- and Y-direction. Selecting a thicker aluminium link would lead to reduced space for the linear actuator. Even so, choosing steel would mean a heavier mechanism. Hence, optimising all the requirements for the prototype developed by us, steel links of 3 mm thickness were used to provide sufficient strength, lower maximum deformation and more space for accommodating the linear actuator. 

The above-mentioned procedure shall be followed to choose various design parameters of the scissor lift mechanism optimally, depending on the size, weight and use case of the robot.

\subsection{Watering and Pesticide Spraying}
\begin{figure}[!tb]
    \centering
    \includegraphics[width=0.9\linewidth]{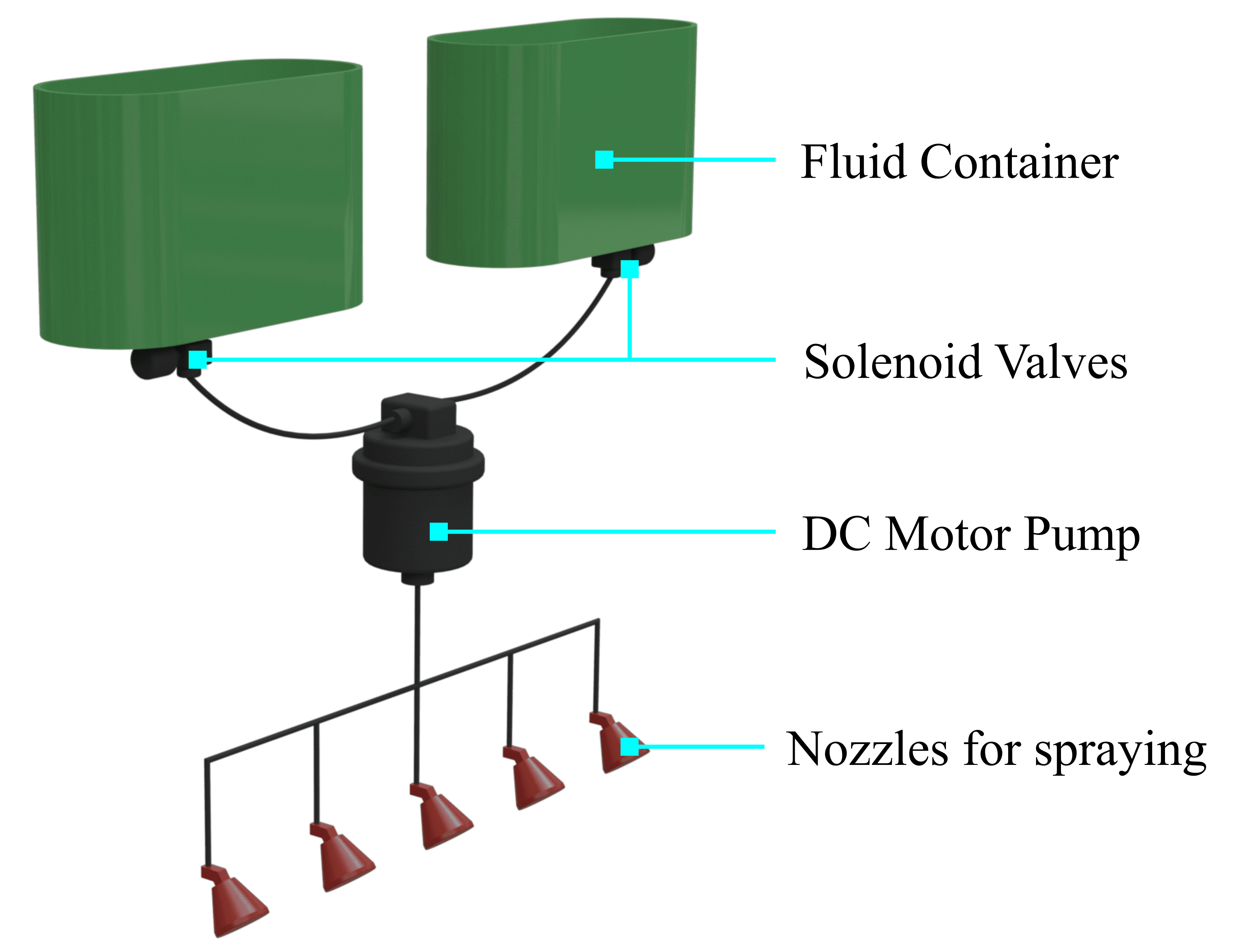}
    \caption{CAD model of Watering and Pesticide Spraying Mechanism}
    \label{fig_watering}
\end{figure}
\begin{figure*}[!tbh]
     \centering
     \includegraphics[width=0.8\textwidth]{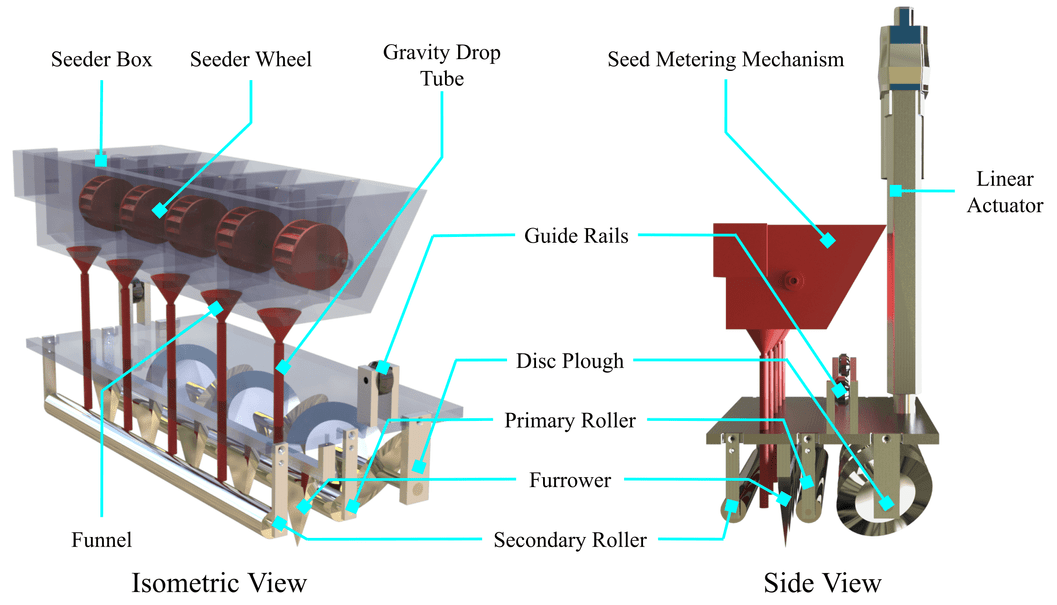}
    \caption{CAD Models of isometric and side views of the ploughing sowing rolling (PSR) mechanism}
    \label{fig_psr}
\end{figure*}
Irrigation and pest control are the two most important farming operations as they directly affect the quality of the crops and hence, the yield.
The prevalent methods for irrigation and pesticide spraying have a lot of drawbacks. Installing systems for drip irrigation, sprinkler irrigation, etc. consumes a significant amount of space and is time-consuming, costly and labour-intensive. There have been multiple attempts to robotise the irrigation process through mobile robots in order to address the drawbacks posed by the traditional irrigation systems in indoor and outdoor environments  \cite{nagaraja2012plant, chen2020design, bodunde2019architectural, adeodu2019development, rafi2016design}. Traditional manual pesticide spraying is highly ineffective and sometimes dangerous. Improper use of pesticides can contaminate groundwater and affect soil fertility \cite{yallappa2017development}. In order to address these issues, several Unmanned Aerial Vehicles (UAVs) have been developed to automate the pesticide spraying process \cite{xue2016develop, huang2015development}. Although it quickens the pesticide spraying, it does not optimise pesticide usage. To solve this problem, mobile robots were then employed to minimise the use of pesticides through controlled spraying. The robot was developed by Geng \textit{et al.} \cite{geng2012assessment} minimized the use of pesticides up to 60\% when compared to the traditional spraying methods. Similar works related to pesticide spraying were proposed in \cite{sammons2005autonomous, oberti2016selective, bechar2016agricultural}. 

All the above-mentioned works provide separate solutions for irrigation and pesticide spraying. We propose a design that provides a combined solution for irrigation and pesticide spraying. The design has been inspired \cite{bhattacharyya2020design} for accommodating more than one type of liquid. When moving through the field, this module allows the robot to spray water or pesticide.

The subsystem consists of storage tanks, pipes, proportional solenoid valves, nozzles, and DC motor-driven pump as shown in Fig. \ref{fig_watering}. There are two storage tanks - one for water and one for pesticide - fixed inside the chassis of the prototype robot. Multiple tanks can be attached to store different chemicals such as insecticides, herbicides, fungicides, etc. depending on the size of the final robot. The solenoid valves are appended below each of the tanks to ensure flow from only one tank at a time. The valves from all tanks are normally kept closed in order to allow flow only when they are actuated. Pipes from these valves extend to the DC motor-driven pump. The pump pressurises the liquid that needs to be sprayed, and also controls the mass flow rate from the nozzles. We use an array of five nozzles, extending from the pump, and placed below the chassis to spray the liquid.

\subsection{Ploughing, Sowing and Rolling (PSR) Mechanism}
A novel mechanism to perform farming operations such as ploughing, sowing, and rolling in series has been developed. The mechanism consists of sub-modules, namely disc plough, a primary roller, furrower, seed metering module and a secondary roller. They are concatenated in series and placed vertically at different heights based upon their respective functions as shown in Fig. \ref{fig_psr}. All these tools are mounted on the base plate which is connected to the chassis through a linear actuator Fig. \ref{fig_psr}. Since this mechanism performs the farming operations sequentially, it reduces the overall work time, total power consumed and cost of operation when performed separately. Moreover, each sub-module is detachable enabling the robot to be used for different types of crops. 

\subsubsection{Disc Plough}
\begin{figure}[!tb]
    \centering
    \begin{subfigure}[b]{0.45\textwidth}
         \centering
         \includegraphics[width=0.9\textwidth]{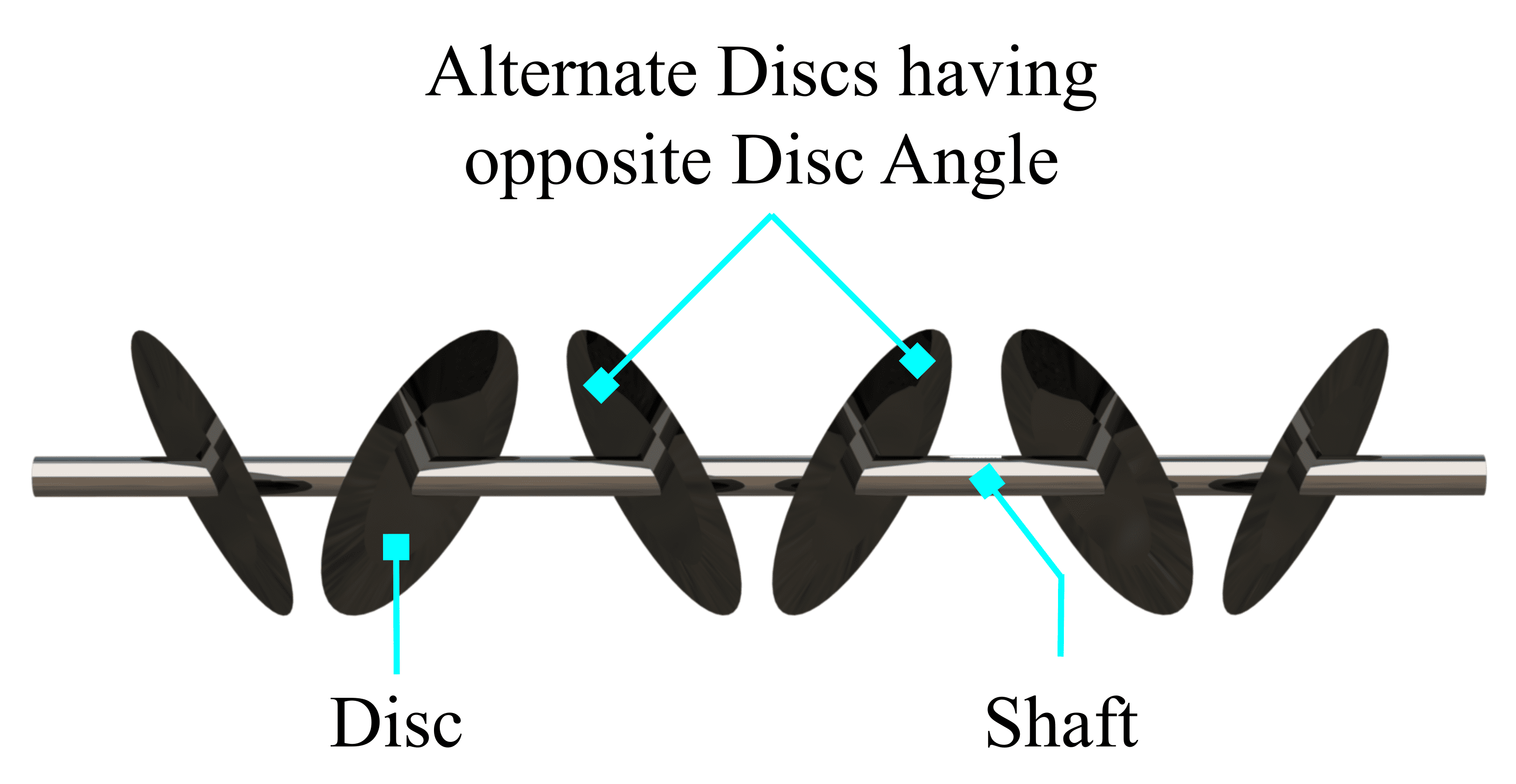}
         \caption{CAD model of the Disc Plough}
         \label{fig_disc_plough}
     \end{subfigure}
     \begin{subfigure}[b]{0.23\textwidth}
         \centering
         \includegraphics[height=3.5cm]{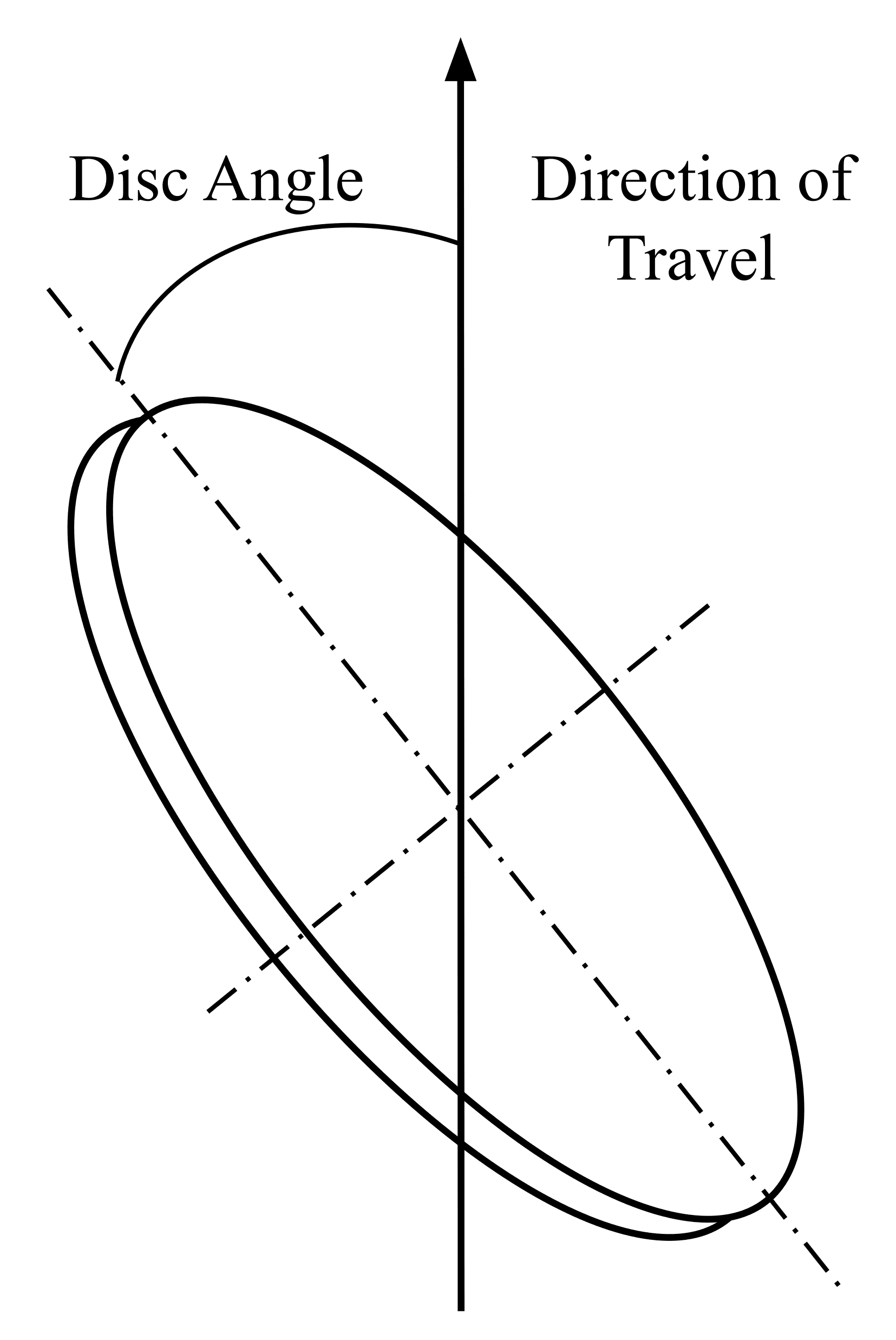}
         \caption{Disc Angle}
         \label{fig_plow_disc_angle}
     \end{subfigure}
     \begin{subfigure}[b]{0.23\textwidth}
         \centering
         \includegraphics[height=3.5cm]{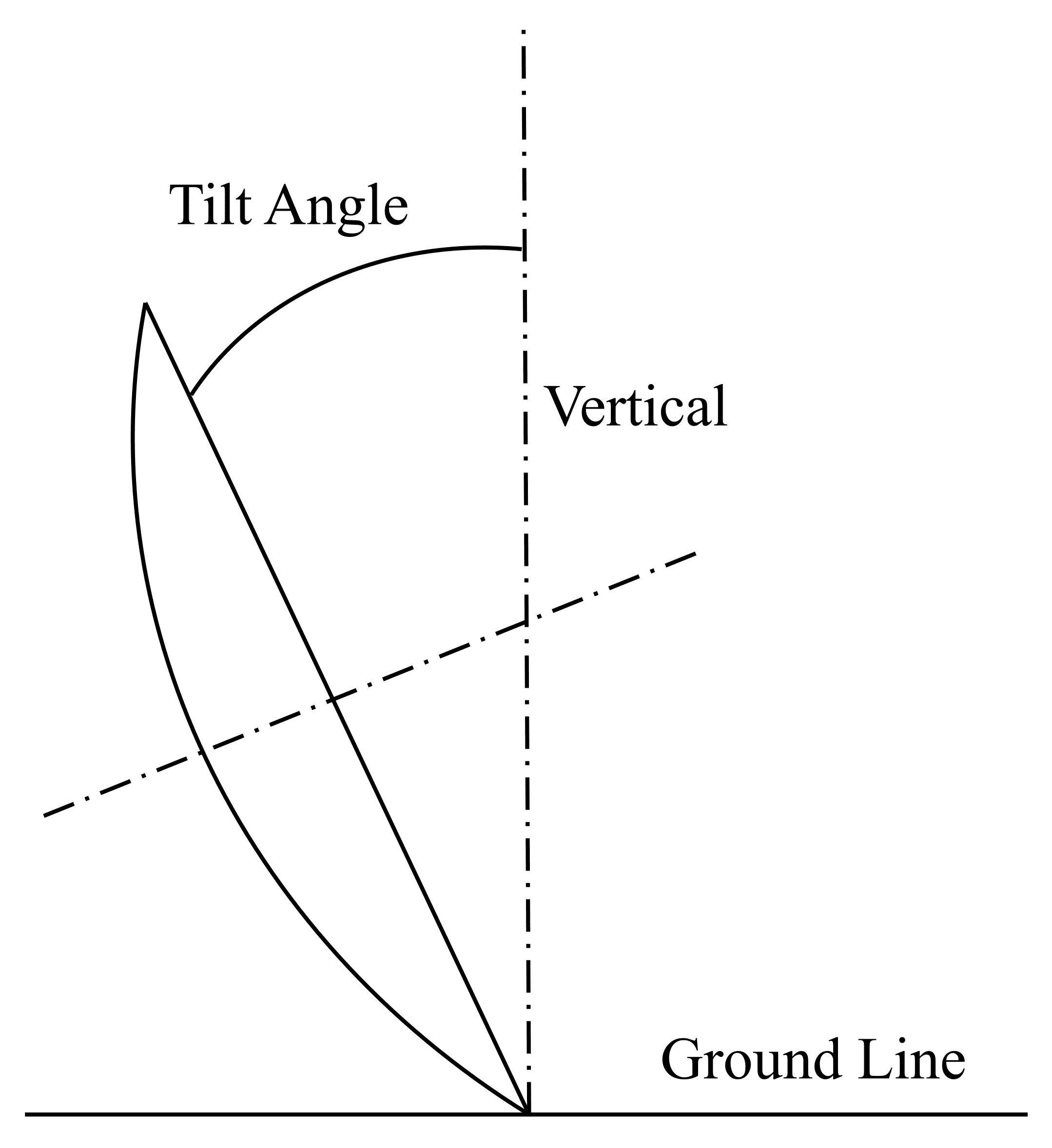}
         \caption{Tilt Angle}
         \label{fig_plow_tilt_angle}
    \end{subfigure}
    \caption{Illustration of the CAD model and disc and tilt angles of the disk plough}
    \label{disc_angles}
\end{figure}
The disc plough attached to the robot performs secondary tillage that helps in seedbed preparations and intercultural operations. It creates refined soil conditions following primary tillage and provides breaking of clods and mixing of crop residues \cite{smith1976farm}. Disc plough is chosen because it is long-lasting, has more mixing action than inversion, and works well in sticky soil conditions \cite{CHANDRAMOULI201823702}. Smooth edged discs have been used as they have low maintenance costs and work well even if substantial parts of the disc are worn out in abrasive soil.

\par The disc angle is defined as the angle at which the plane of cutting edge of the disc is inclined to the direction of travel, as shown in Fig. \ref{fig_plow_disc_angle}. The tilt angle is defined as the angle at which the plane of cutting edge of the disc is inclined to vertical, as illustrated in Fig. \ref{fig_plow_tilt_angle}. In \cite{kepner1990principle}, it was found out that these angles influence the depth and width of cut of the soil. It was concluded in \cite{osman2011effects} and \cite{bukhari1992effect} that, with an increase in disc and tilt angles, the field capacity and the power required to plough the field increased. Hence, it becomes important to choose suitable values for these angles while designing the mechanism. From \cite{article_el_naim}, the optimum values of disc and tilt angles are found to be between $42^\circ$ to $45^\circ$ and $15^\circ$ to $20^\circ$.
\par In Aarohi, six circular smooth edge discs attached to a shaft have been used. They are oriented with disc angle and tilt angle as $45^\circ$ and $15^\circ$, respectively, in order to reduce the radial force. Alternative discs are oriented in opposite directions to eliminate side thrust force on the robot. This novelty in design has allowed removing the additional rear furrow wheel which is used in the reduction of side thrust in conventional mechanisms.
The shaft that connects the discs is powered by a motor to control the rotational speed of the discs. The shaft is rotated such that the friction acts on the discs in the direction that supports the motion of the robot reducing the load on the motors that power the wheels.

\subsubsection{Primary Roller}

It is placed immediately after the disc plough with sufficient clearance. The primary roller levels the ploughed soil, thereby flattening the soil. Torsional springs are mounted at the ends of the roller to give compliance and reduce additional loads on the wheels of the robot and ensure proper levelling of the soil.

\subsubsection{Furrower}
V-shaped furrowers are used to make grooves of appropriate depth on the flattened land. In the robot, four V-shaped metal plates are connected to the PSR mechanism’s base plate through a bracket. Although the V-shaped metal plates are not the most optimal design for furrowers \cite{smith1976farm}, they have been considered due to their relative simplicity in manufacturing and assembly. The linear actuator connected to the base plate provides the downward force required for furrowing. 

\subsubsection{Seed Metering Module}
\begin{figure}[!tb]
    \centering
     \includegraphics[width=0.45\textwidth]{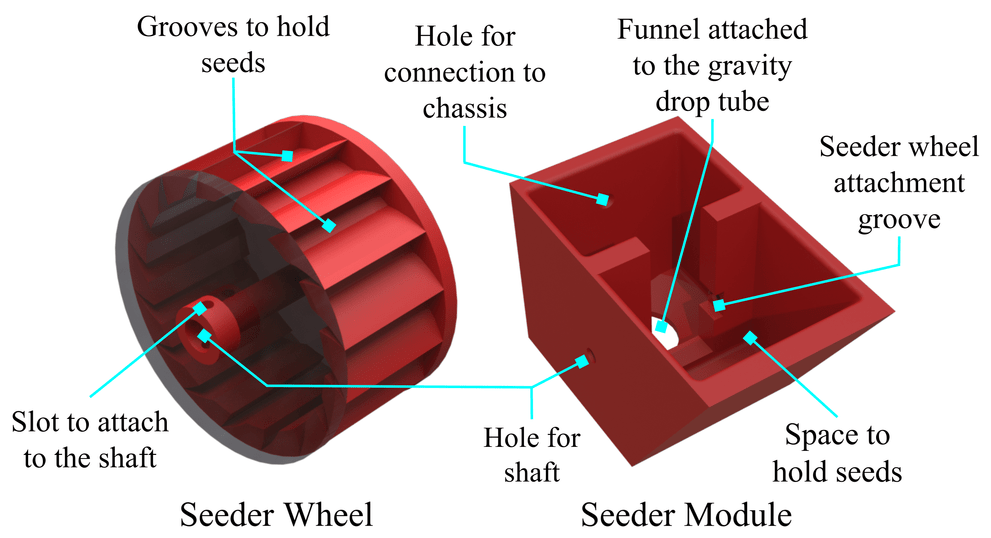}
     \label{fig_seeder_module}
    \caption{CAD Models of the seeder wheel and the seeder box of the seed metering module}
    \label{fig_seeder}
\end{figure}
\begin{figure*}[!tb]
    \centering
    \includegraphics[width=0.85\linewidth]{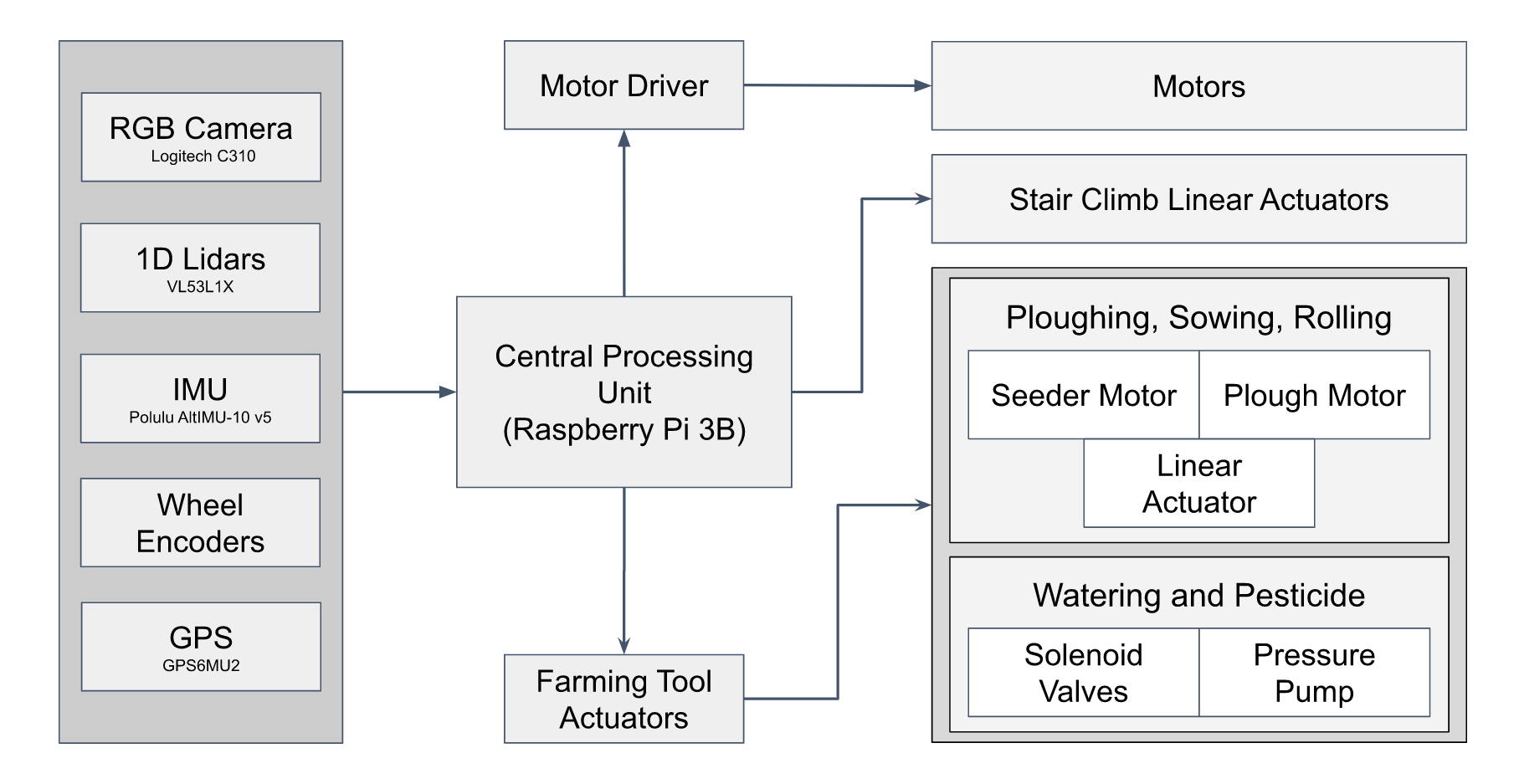}
    \caption{Sensing and Interfacing}
    \label{fig_Sense}
\end{figure*}
Sowing seeds precisely is the next and a very crucial step in precision agriculture. Recently, a lot of researchers have been focussing on precision seeding techniques as a driving direction for modern agriculture technology \cite{singh2005optimisation, yasir2012design}. It can save a lot of seeds and effectively control the inter-seed distance as well. Broadly, two kinds of seed metering devices for grains have been proposed in literature based on the mode of operation: (a) Mechanical seeder, and (b) Pneumatic seeder. While pneumatic seed metering devices have proven to have better performance in terms of a more uniform distribution and quality of feed index, several researchers have advocated the use of mechanical seeding for its low cost and reliability especially for the sowing of crops like wheat \cite{minfeng2018optimal, yu2021design} . In conventional mechanical feeders, a ground wheel is typically used to drive the seed metering device \cite{article_ground_wheel}. However, occasional wheel slips and resistance in an uneven terrain results in poor performance of this system in controlling the seed rate. Zhai \textit{et al.} \cite{jianbo2014design} improved upon this system by designing a control system that could keep the rotational speed of the seed-metering device consistent with the seeder’s working speed and achieved good performance. Jin \textit{et al.}  \cite{minfeng2018optimal} also solved the problem of poor consistency management by adjusting the opening of fluted rollers by a novel mechanism. Our seeder system is a low-cost yet efficient solution to the automated seeding process and drives inspiration from these designs.

As shown in Fig. \ref{fig_seeder}, the seed metering module in Aarohi consists of four boxes from where a seed feeder wheel puts the seeds in the funnel, which is then connected to a gravity drop seed tube. The other end of the tube is placed just behind the V-shaped metal plates of the liner, to ensure the falling of seeds inside the furrow. The seeder boxes are slotted and attached to the chassis. The seed feeder wheels are connected to a shaft which is in turn connected to a stepper motor. This stepper motor is attached to the chassis through L-brackets and it is responsible for the prevention of interlocking among the seed granules and also controlling the longitudinal inter-seed distance during sowing operation.

\subsubsection{Secondary Roller}
A secondary roller is used to fill the furrows and level the ground. This rolling operation creates a smooth and firm seedbed, presses the seeds into the soil thereby eliminating air pockets, and helps in retaining the soil moisture for a longer period, thereby allowing for faster germination \cite{smith1976farm}.

\begin{table*}[!tbh]
    \centering
    \small
    \caption{Description of the count and locations of 1D LiDAR sensors. Here $N$ is the number of wheel pairs in the robot, and since the current prototype has two-wheel pairs $N=2$}
    \begin{tabular}{ c c c }
        \toprule
        \textbf{\centering Location} & \textbf{\centering Count} & \textbf{\centering Symbols}\\
        \midrule
        At far ends of all sides  & 2 per side x 4 sides = 8 & $(L_{fl}, L_{fr}), (L_{bl}, L_{br}),  (L_{lf}, L_{lb}),  (L_{rf}, L_{rb})$\\
        Downward Facing & $N$($=2$ scissors) + 2(chassis) = 4 & $(L_{d1}, \hdots, L_{dN})$, $(L_{df}, L_{db})$\\
        Width Estimation (at $45^\circ$) & 2 (one for each side) & $(W_{l}, W_{r})$\\
        Failsafe (at $25^\circ$) & 1 per side x 4 sides = 4 & $(F_{f}, F_{b}, F_{l}, F_{r})$\\
        \bottomrule
    \end{tabular}
    \label{tab:lid_table}
\end{table*}
\begin{figure*}[!tbh]
    \centering
    \includegraphics[width=0.9\linewidth]{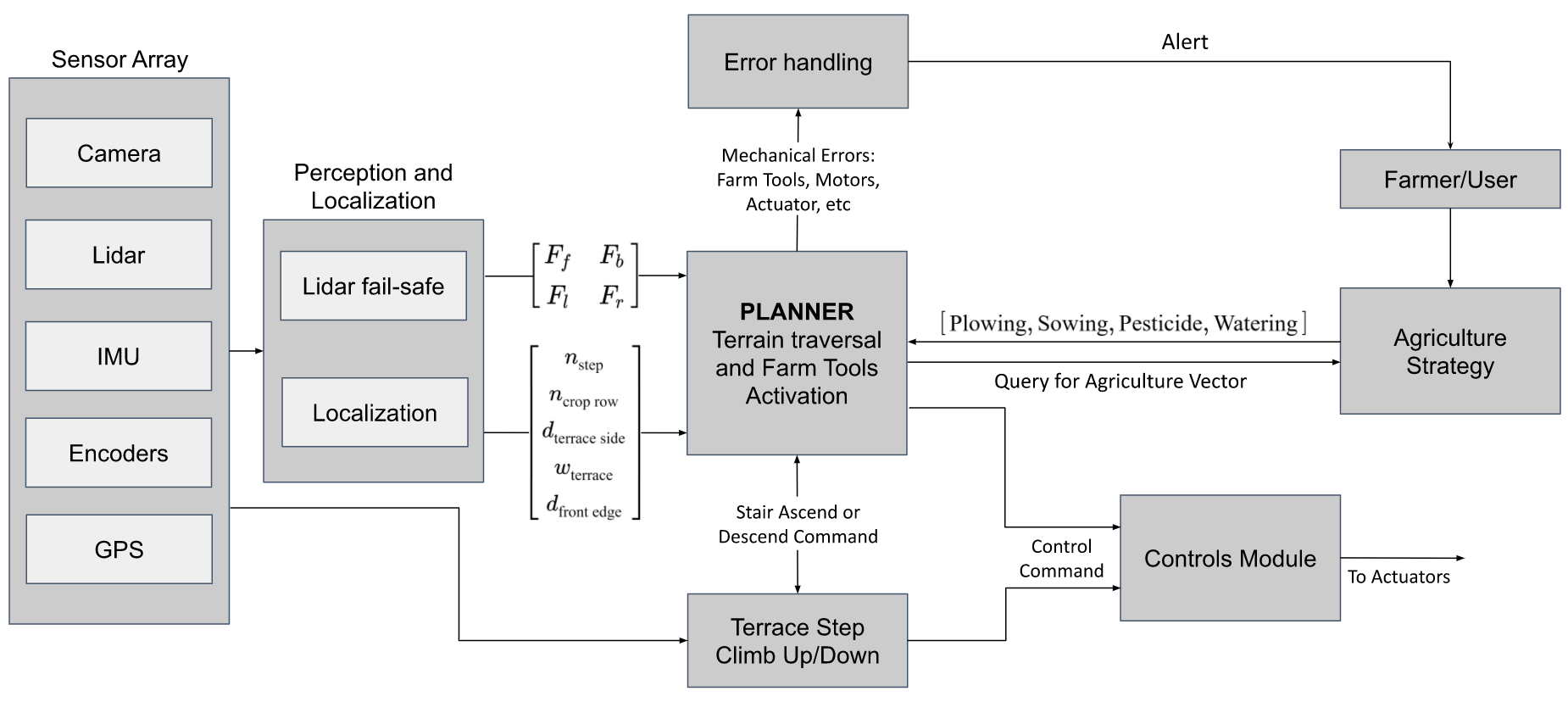}
    \caption{Software Architecture Overview}
    \label{fig_software_flowchart}
\end{figure*}

\section{Software Architecture}
\label{Software_Architecture}
A modular Robot Operating System (ROS) based software stack is developed to achieve fully autonomous terrace navigation and farming, as shown in Fig. \ref{fig_software_flowchart}. We first discuss the localisation module, which uses a combination of 1D lidars, RGB camera, GPS, IMU  and wheel encoders (as shown in Fig. \ref{fig_Sense}) to localize the robot within its environment and estimate terrace edge widths. Next, we discuss the planning module that outputs the desired high-level plan to traverse the terrain and perform the given farming operations. Finally, we discuss the controls module that calculates actuator inputs based on the generated high-level plan, while accounting for any external disturbances and uncertainties.

\subsection{Perception and Localisation}
\begin{figure}[!tb]
    \centering
    \includegraphics[width=0.60\linewidth]{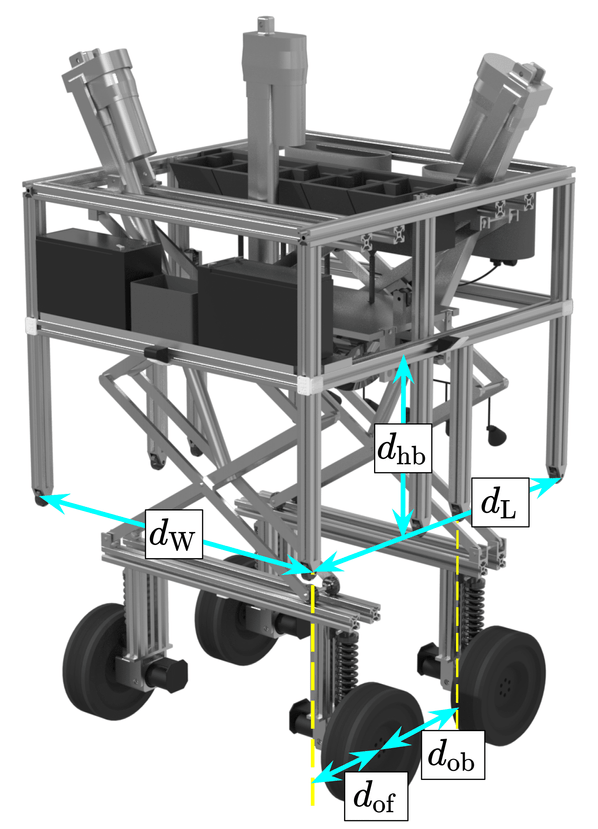}
    \caption{Illustration of various robot lengths and respective notation}
    \label{fig_robot_lenghts}
\end{figure}
\begin{figure}[!tb]
    \centering
    \includegraphics[width=0.9\linewidth]{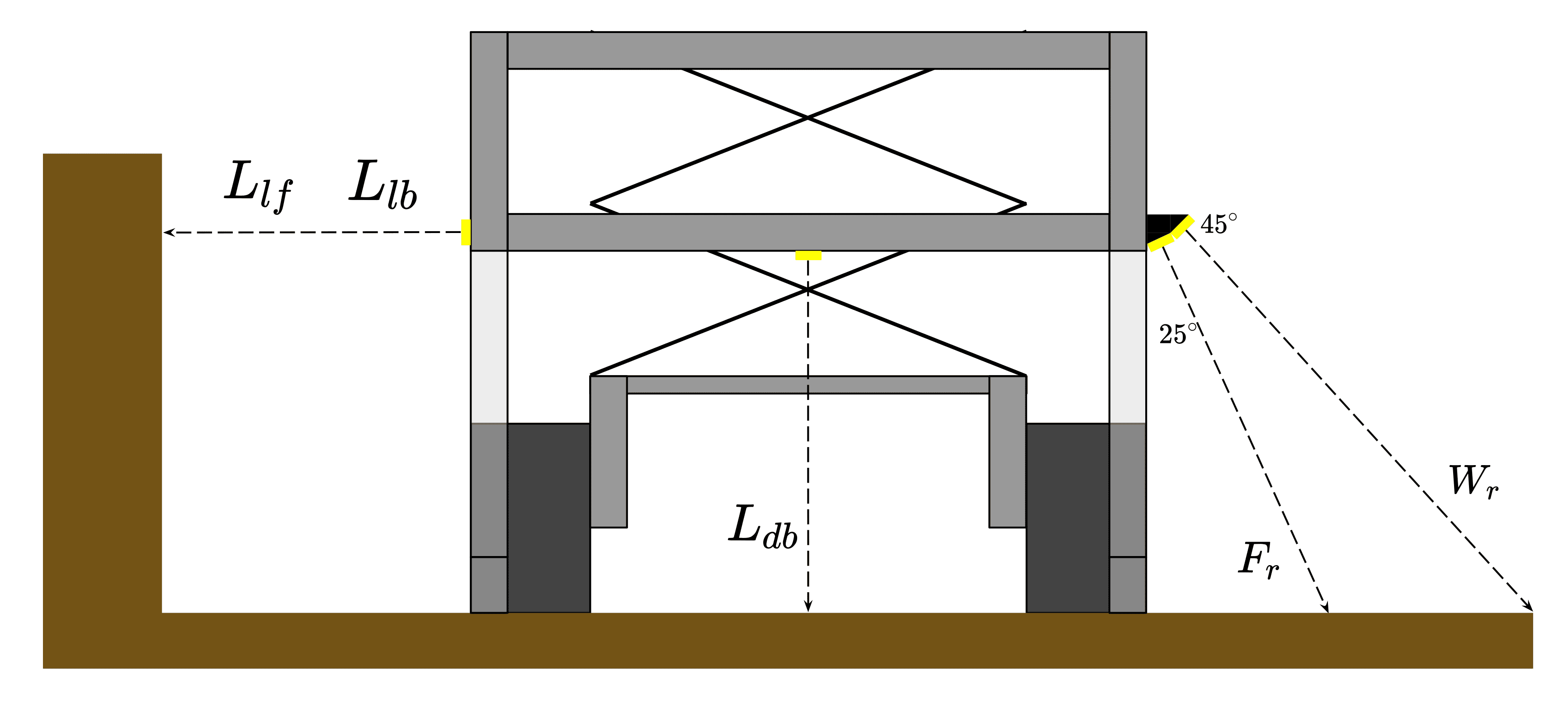}
    \caption{Illustration for width estimation}
    \label{fig_width_estimation}
\end{figure}

\subsubsection{Localisation}
The estimation of the pose and orientation of the robot is done using ground odometry data from wheel encoders and inertial measurements from IMU. As ground odometry tends to be inaccurate in the presence of wheel slipping, it is often fused with other forms of inputs like lidars and GPS for accurate localisation. We use an Extended Kalman Filter (EKF) method to fuse different localisation estimates. While GPS sensors provide global position data, they were not found to be of much use in the robot due to large errors (\~1m). The EKF method effectively weights the different estimates with the uncertainty associated with them and outputs the robot position. Based on these low-level inputs, the overall localisation module as shown in Fig. \ref{fig_software_flowchart}, outputs high-level location information that includes the step number (from the top), crop row number and distance from sides of the terrace. 

\par We use the width estimation lidars $W_l$ and $W_r$ to estimate the minimum terrace width at the current step. These lidars are pointed downwards at an angle of $45^\circ$ from the horizontal. For the purpose of describing the method for terrace width estimation, assume that the robot is moving such that the terrace wall is towards its left, as shown in Fig. \ref{fig_width_estimation}. A new width measurement is taken when there is a sudden change in the readings of lidars $W_l$ and $W_r$. Using simple geometry, the terrace step width can be estimated as
\begin{equation}
    w_{\rm terrace} = \left[\frac{L_{lf}+L_{lb}}{2} + d_W+ L_{db}\tan{45^\circ}\right]\cos\phi
\end{equation}
Where $\phi$ is the angle between robot heading and the terrace wall, given by
\begin{equation}
    \phi= \tan^{-1}\left({\frac{L_{lf}-L_{lb}}{d_L}}\right)
\end{equation}
Here $d_W$ and $d_L$ are the robot width and lengths, as illustrated in Fig. \ref{fig_robot_lenghts}. This width estimation system allows for the robot to handle terrace steps with variable widths along the length, and allows the planning module to account for the variations accordingly. Since the system depends on sudden changes in the width estimation lidar readings, it can estimate width only when the nearest distance between robot midline and the terrace step end crosses the value $L_{db}\tan 45^\circ$.

\par
\begin{figure}[!tb]
    \centering
    \includegraphics[width=0.9\linewidth]{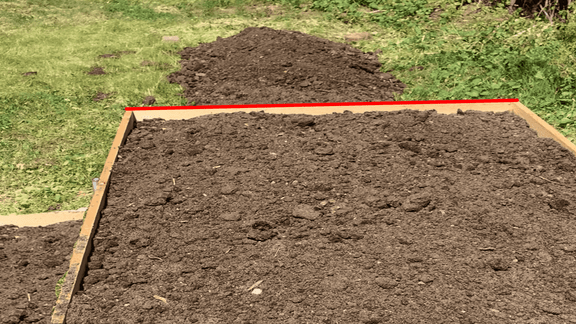}
    \caption{Terrace Edge Detection}
    \label{fig_edge}
\end{figure}

 To measure the distance from the robot's body-fixed frame and the terrace edge, an image processing based edge detection algorithm is deployed. It uses a modified urban road stop-line detection algorithm as proposed in \cite{seo2014vision}. A vertical gradient filter is applied on the input image which highlights horizontal features. Next, the probabilistic hough transform is applied on sufficiently large clusters with a horizontal spread to detect horizontal lines, and relatively close horizontal lines are merged into super-clusters, and the bottom-most super-cluster is detected as the terrace edge, as shown in Fig. \ref{fig_edge}.
Inverse homography and static frame transforms are used to convert the distance pixels to meters and calculate the distance between the terrace edge and the robot's body-fixed frame. We test and validate the algorithm multiple times on even and uneven edges to ensure robust detection. 

\subsubsection{Fail-safes for fall avoidance}
The fail-safe system for fall avoidance comprises of four 1D lidars mounted at all the edge centres of the robot, looking downwards making an angle $25^\circ$ with the horizontal Fig. \ref{fig_width_estimation}. These lidars form a safe zone around the robot, and if a sudden increase in values of any of these lidars is detected, an emergency stop is triggered and all autonomous operations are stopped. The safe zone is an imaginary rectangle with its edges located at an inner distance $L_{db}\tan 45^\circ$ from the robot edges, which, when the linear actuators are fully contracted, is about 12 centimetres. This system stops the robot from falling off terrace edges, which might happen due to planning errors or system uncertainties.

\begin{figure*}[!tbh]
    \centering
    \includegraphics[width=0.88 \textwidth]{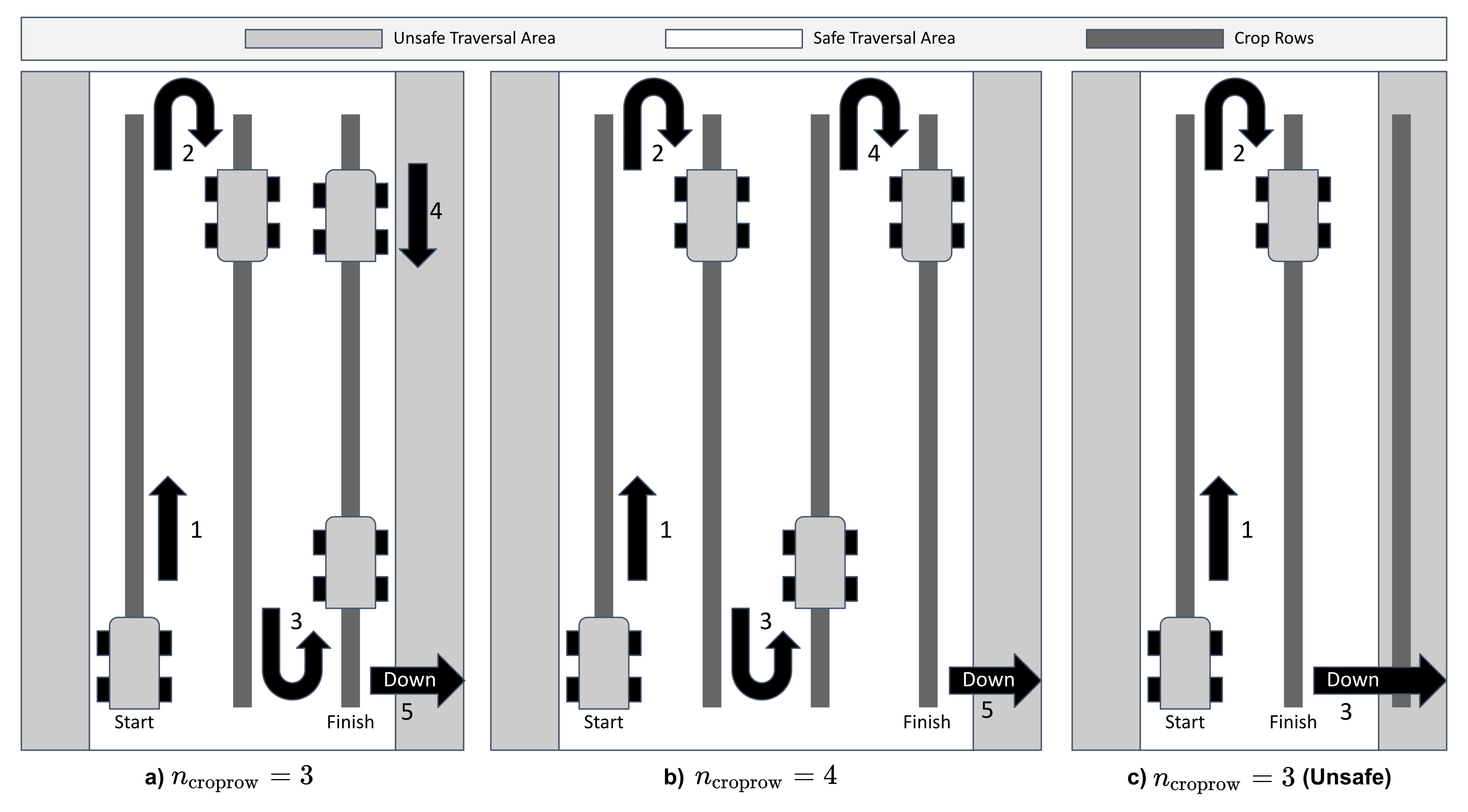}
    \caption{Traversal along Terrain}
    \label{nav_plan}
\end{figure*}
\label{subsec_stair_climb}
\begin{figure*}[!tbh]
    \centering
    \includegraphics[width=0.8\textwidth]{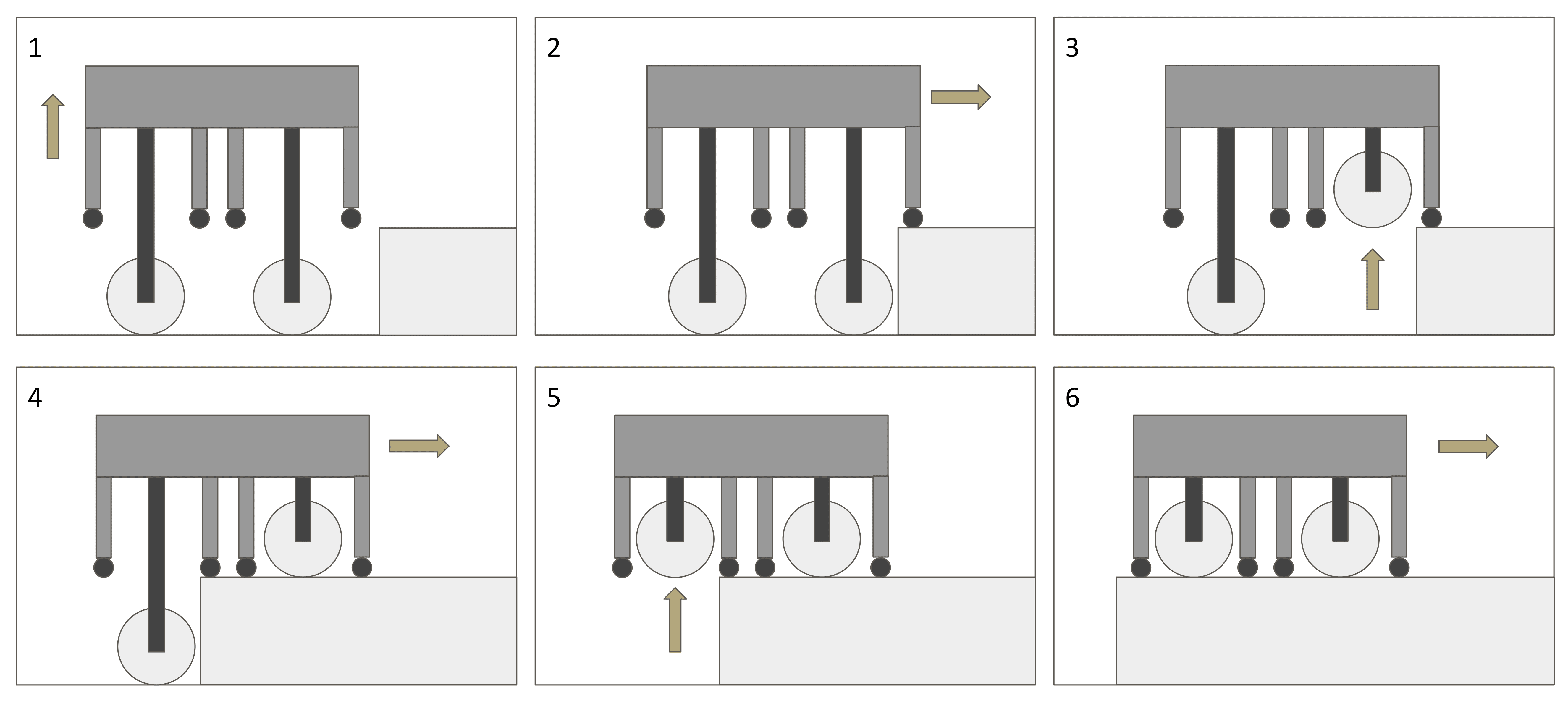}
    \caption{Stair Climbing Algorithm Illustration}
    \label{stair_climbing}
\end{figure*}
\subsection{Planner}
\subsubsection{Row Traversal}
Based on the number of rows of crops present in the terrace, the planner generates the primary path that the robot has to traverse on the terrace. Within the current method, the primary path is a straight line along the terrain at a constant distance from the wall of the terrace. The planner takes as input the number of rows of crops to devise an optimal path that traverses through each plant row while minimising time and actuation cost. As exemplified in Fig. \ref{nav_plan}, if $n_{\rm crop row} = 3$, the path generated ensures that the robot returns in the same orientation as it entered in order to save actuation costs of zero degrees turn. The planner also takes into account the spatial constraints of the robot and the terrace width estimate to maintain safe margins from both sides of the terrace. The planner prevents the robot from entering into any unsafe zones along the terrace.

\subsubsection{Agriculture Strategy}
The software stack provides the provision for the farmers to input a desired agricultural strategy which acts as an input to the planner. Before starting the work on terrain, the planner sends a query to the agriculture strategy module as shown in Fig. \ref{fig_software_flowchart}. Based on the user inputs provided, the planner module accordingly plans the traversal path as well as the activation required for different farming tools to perform agricultural activities. Hence a flexible strategy can be maintained from terrain to terrain and crop to crop. Different combinations of farming activities can be fed into the strategy module in order to achieve different goals under varying circumstances.

\subsection{Stair Climbing and Descending Module}
On completion of all the tasks on the respective step, the planner activates the stair climbing/descending module. As given in Fig. \ref{fig_software_flowchart}, the planner sends a high-level command for performing the climb up or climb down activity. Before the actual stair climbing or descending operation, the robot is aligned parallel to the stairs. Using the pair of lidars facing the wall, the robot aligns itself parallel to the wall by using a yaw control algorithm that provides actuation inputs according to the differential value of distances as measured by both the lidar. Once the robot is aligned, the stair climb/descend algorithm takes over. 
\begin{algorithm2e}[!tb]
\caption{Climb Up}
    {$ N = \text{Number of wheel pairs} $}\\
    {$yawAlignment()$}    \\
    \While{$! DetectSuddenChange(L_{fl,fr}.distance)$}
    {
  	    ExtendAllScissors()
    }
  {$stepHeight = L_{df}.distance + d_{hb}$}\\
  \While{$L_{df}.distance \leq StepHeight$} 
    {
  	    ExtendAllScissors()
    }
  \While{$L_{df}.distance \geq StepHeight$}
    {
  	    MoveForward()
    }
    \For{$i\gets1$ \KwTo $N$ }
    {
        \While{$L_{di}.distance \leq Step Height$} 
        {
      	    CompressScissor($i^{th}$ pair)
        }
        $targetDistance = L_{fl,fr}.distance - ( d_{of} + d_{ob} )$\\
        \While{$L_{fl,fr}.distance \geq targetDistance$} 
        {
      	    MoveForward()
        }
    }
\end{algorithm2e}
\par As seen in Fig. \ref{stair_climbing}, while climbing, initially the robot is at its bottom-most position. In the first stage, the scissor mechanism lifts the robot to the height of the step. In the second stage, the  moves forward such that the front dummy wheels are in contact with the horizontal surface of the next step. Here we use LiDAR $L_{df}$  to detect the edge. In the next stage, the front pair of wheels are lifted up by the scissor mechanism while the rear wheels are still in contact with the present step. In the fourth stage, the robot moves forward such that the front wheels and the third pair of dummy wheels come in contact with the next step. Here we use LiDAR $L_{fr}$ and $L_{fl}$ to ensure precise movement. In the fifth stage, the rear wheels are lifted up and in the last stage, the robot moves forward such that the whole robot finally moves on to the next step. Fig. \ref{stair_climbing} illustrates the above-explained stages. In every stage, it is ensured that the centre of mass of the whole robot stays inside the polygon made by the contact points of the wheels and the surface to ensure its stability. For climbing down, the same stages would be followed in the reverse direction. 

\begin{algorithm2e}[!tb]
\caption{Climb Down}
    {$ N = \text{Number of wheel pairs} $}\\
    {$yawAlignment()$}\\
    \For{$i\gets1$ \KwTo $N$ }
    {$targetDistance = L_{bl,br}.distance + d_{of} + d_{ob}$
       \While{$L_{bl,br}.distance \leq targetDistance$} 
       {
       	    MoveForward()
       }
       \While{$L_{di}.distance \geq d_{hb}$} 
       {
       	    ExtendScissor($i^{th}$ pair)
       }
    }
    $targetDistance = L_{bl,br}.distance + d_{of}$ \\
    \While{$L_{bl,br}.distance \leq targetDistance$} 
    {
           MoveForward()
    }
    \While{$L_{db}.distance \geq d_{hb}$} 
    {
        CompressAllScissors()
    }
\end{algorithm2e}
\par The scissor mechanism is actuated using a linear actuator fixed at orientation with optimised stroke length and force during the process. The mechanism is stable in lifting up to 50 kgs in our robot. It also enables intermediate height control for the robot within 40 cm which is useful for variable height steps in real-life situations. It’s advantageous during the watering and harvesting of crops of different heights. Lastly,  the mechanism dimensions can be modified easily for higher steps.

\subsection{Controls Module}
Using the task strategy and the trajectories from the high-level planner, the control module provides the required actuator inputs for realising the plans as shown in Fig. \ref{fig_software_flowchart}. The overall control module for the robot is solves two broad sub-problems. First is controlling the height of the robot and scissor mechanism actuated with linear actuators and the second is to accurately track a path on the uneven terrain as a differential drive robot. Considering that the control strategy for each of these sub-problems are not correlated, we treat both problems independently and show the controller performance and system stability for each of the sub-problem in this section.
\subsubsection{Pitch Control}
\begin{figure}[!tb]
    \centering
    \includegraphics[width=7.3cm]{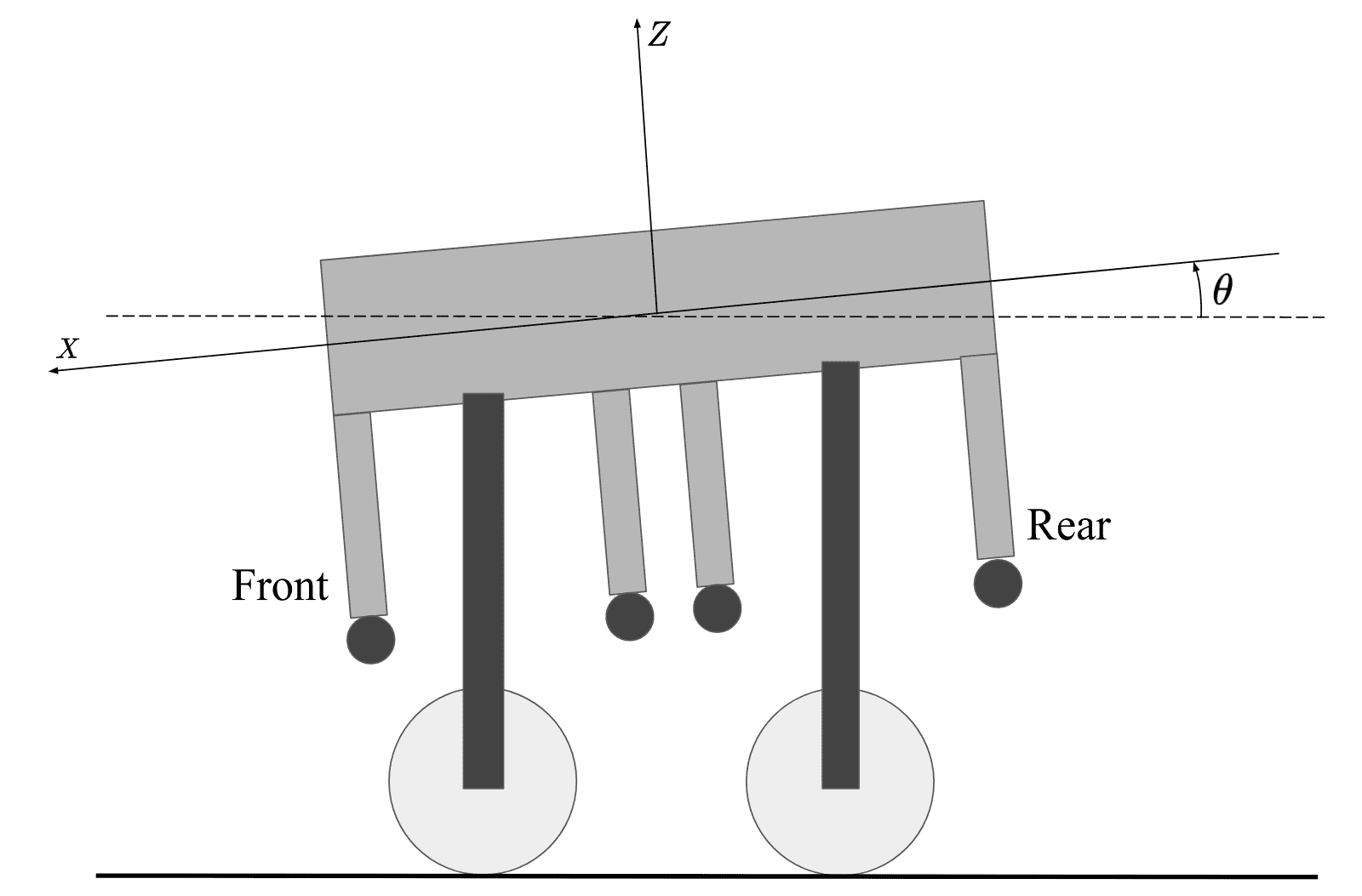}
    \caption{Illustration for pitch control of the robot}
    \label{fig_pitch}
\end{figure}
Due to the difference in the mechanical response of both scissor lift mechanisms, the pitch of the chassis while lifting up may not be zero when both the actuators are given the same input causing instability. We use a PID controller for maintaining zero pitch that ensures the robot remains stable at all times. We estimate the error in pitch and provide corrective feedback to the linear actuators to keep the pitch zero at all times.
\begin{subequations}
\begin{align}
    A_f - A_b =& k_p \theta + k_i \int_{}^{} \theta dt + k_d  \frac{d\theta}{dt} \\
    A_f + A_b =& c_1 - c_2 \Delta v
\end{align}
\end{subequations}
Here $A_{f}$ and $A_{b}$ actuator inputs for the front and back linear actuators. $\theta$ denotes the pitch of the robot, as shown in Fig. \ref{fig_pitch}. This method allows us to achieve two different objectives while controlling the pitch of the robot,
\begin{enumerate} 
    \item Difference in input is controlled based on the pitch of the chassis.
    \item Average actuation input is controlled based on the change in velocity of the base to ensure a constant velocity of ascent/descent.
\end{enumerate}


\begin{figure}[!tbh]
    \centering
    \includegraphics[width = 0.45 \textwidth]{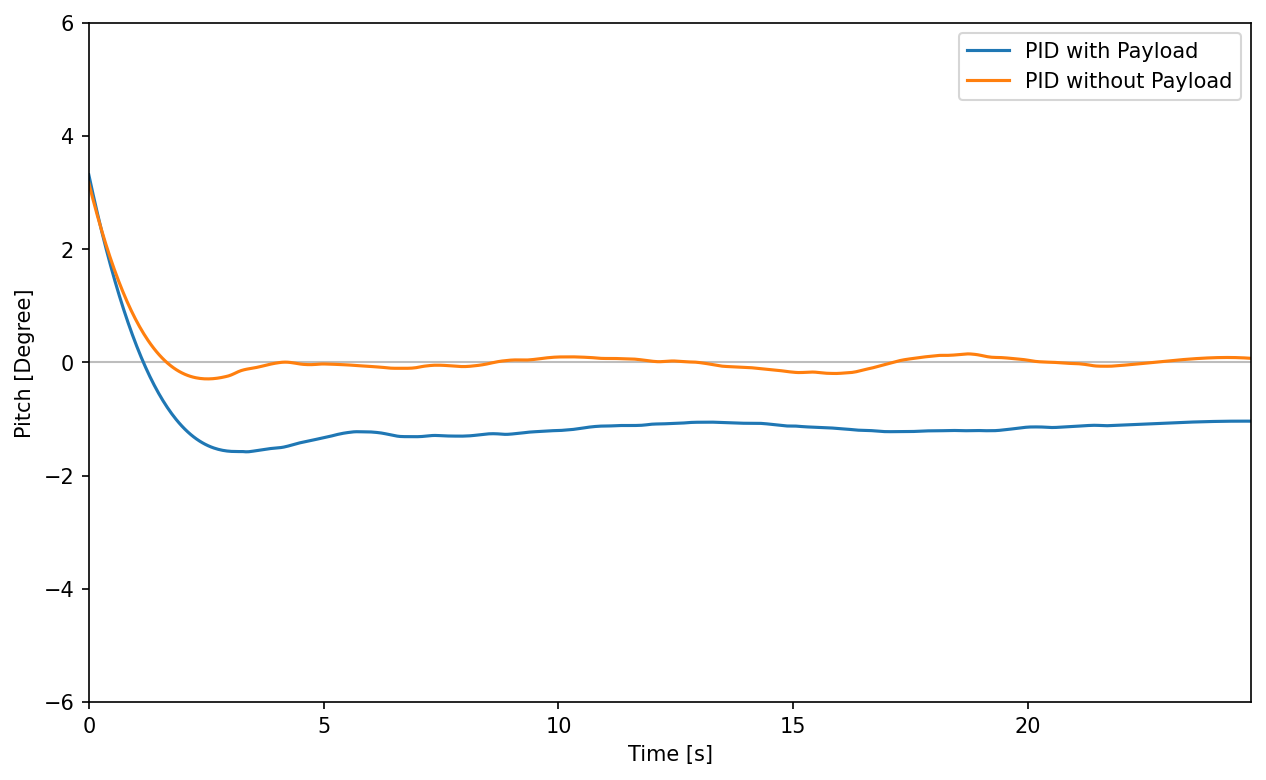}
    \caption{Performance of standard PID controller}
    \label{fig:pid}
\end{figure}

We observe that a standard PID performs well during normal operation but the controller performance depletes on addition of a payload and in presence of external forces. As observed in Fig. \ref{fig:pid}, a standard PID in presence of an external weight leads to a change in the asymptotic value of the pitch which can cause instability. In order to address this we use an adaptive PID algorithm based on \citep{Alonso2013SelftuningPC}. The algorithm uses a modified version of the MIT rule for altering PID gains given as,

\begin{subequations}
\begin{align}
    & k_p(t+1) = k_p(t) + \gamma_{P}\cdot[e(t)-e_m(t)] \\
    & k_i(t+1) = k_i(t) + \gamma_{I}\cdot e_m(t) \\
    & k_d(t+1) = k_d(t) + \gamma_{D} \cdot [\Dot{e}(t)-\Dot{e}_{m}(t)]
\end{align}
\end{subequations}
where $ \gamma_{p}, \gamma_{I}, \gamma_{D}$ are tunable parameters and $e(t),\Dot{e}$ are standard error and error rate measurements whereas $e_{m}(t),\Dot{e}_{m}(t)$ are filtered values through a low pass filter. This algorithm helps to handle noises in the system effectively while preventing extremely high values for the PID gains. 
\begin{figure}[!tbh]
    \centering
    \includegraphics[width = 0.45 \textwidth]{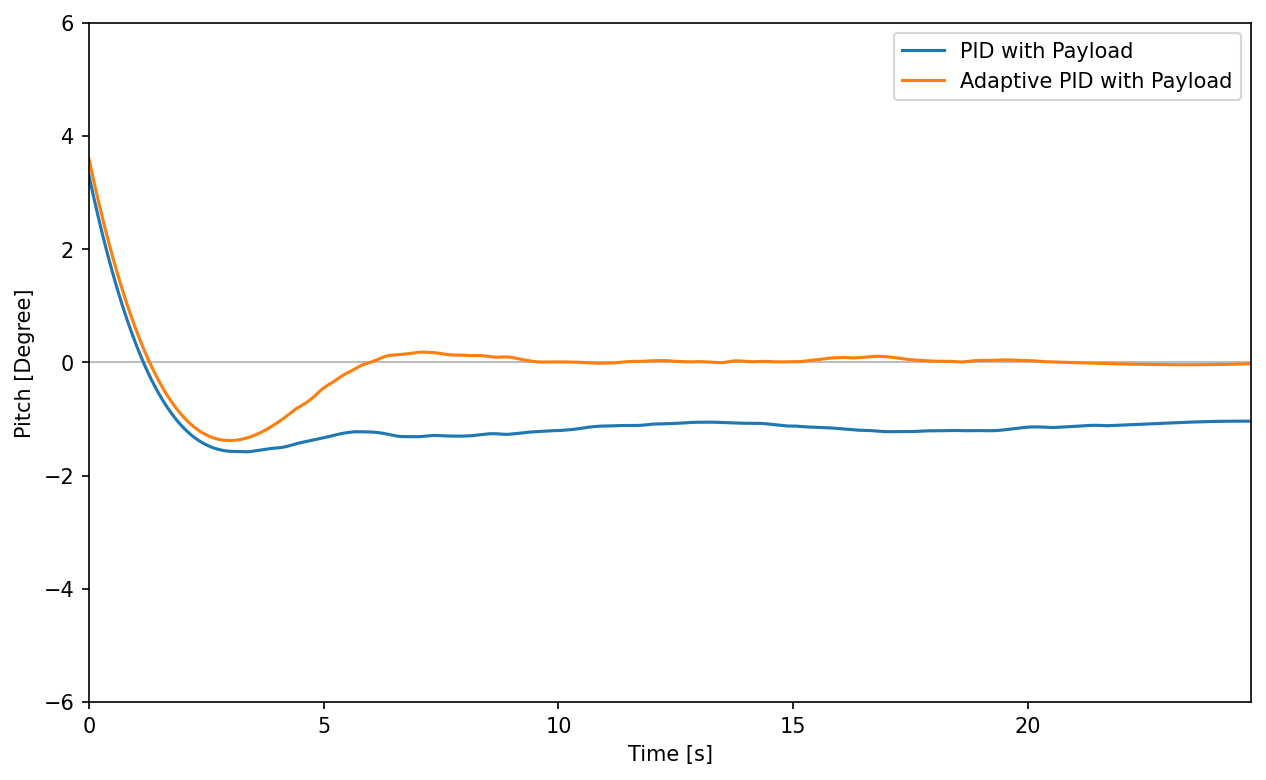}
    \caption{Performance of Adaptive PID controller}
    \label{fig:apid}
\end{figure}
As shown in Fig.\ref{fig:apid}. We observe improved performance with adaptive PID in comparison to standard PID controller in maintaining zero pitch with external disturbances.

\subsubsection{Path Tracking Control}
\begin{figure}[!tbh]
    \centering
    \includegraphics[width=0.35 \textwidth]{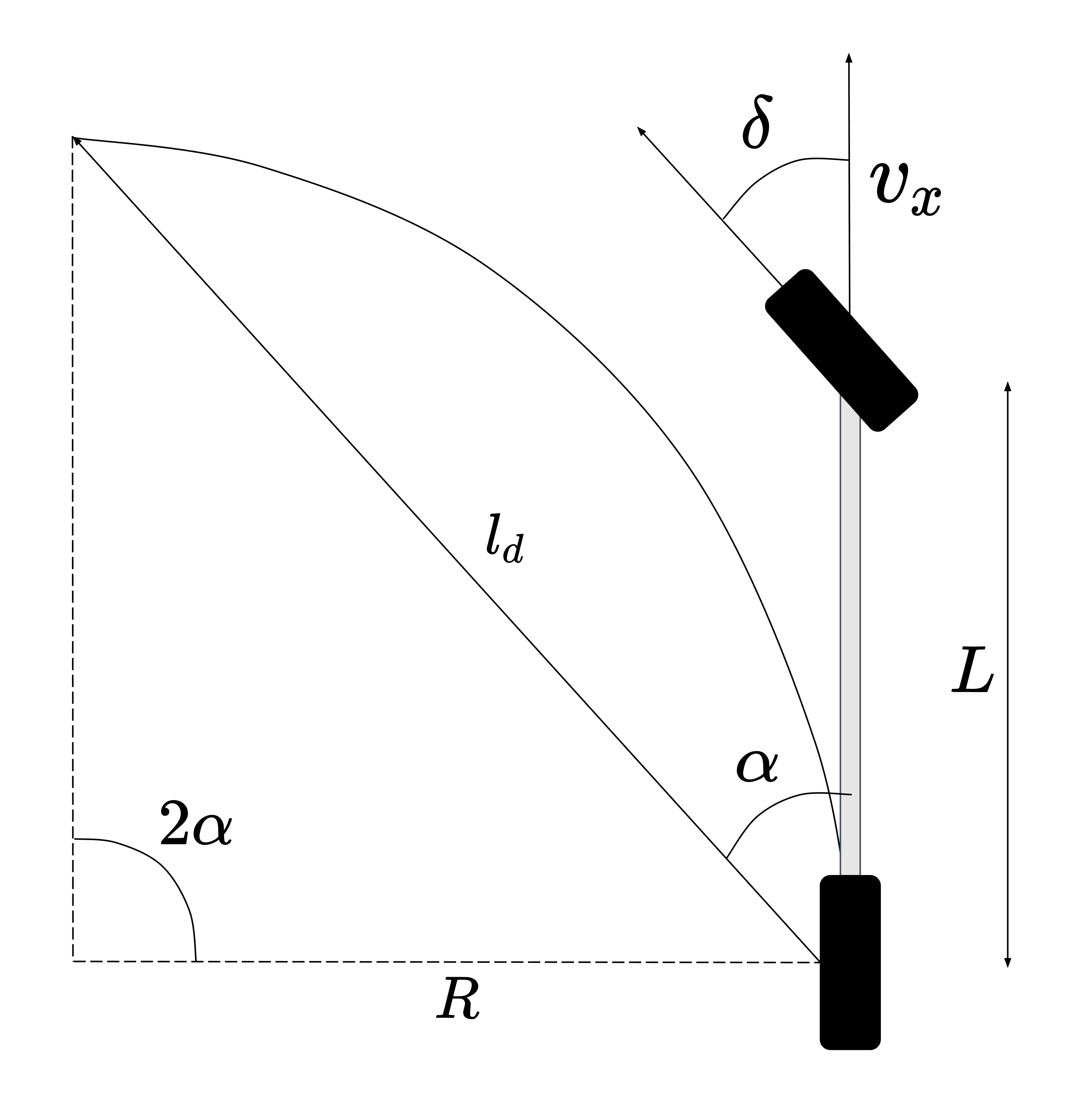}
    \caption{Pure Pursuit Geometry}
    \label{pure_pursuit}
\end{figure}
Given an optimal path from the planning module, we need to ensure accurate tracking of the path considering external forces and environment uncertainties. We use a modified pure-pursuit based controller for this purpose which takes into consideration the kinematic constraints of the robot and external disturbances due to uneven terrains. Pure pursuit is a goal-oriented approach where the wheel velocities are calculated to drive the robot along a circular path from the current position to a final goal position on the optimal path which is at a particular look-ahead distance from the robot. As shown in Fig. \ref{pure_pursuit}, the robot considers the point on the path at a distance of $l_{d}$ as the goal. For smooth trajectories, the robot is treated as an ackermann vehicle, and based on Ackermann geometry, $\delta$ is the steering angle, $v_{x}$ is the longitudinal velocity and $R$ is the turning radius. $\alpha$ is the central angle for the turn and its value is calculated from the differential value of the distance measured by the pair of lidars facing the adjacent wall.
The look-ahead distance is the parameter that needs to be
tuned. We use an adaptive look-ahead distance proportional to the velocity of the robot to ensure smooth control inputs. Hence, we obtain the required steering angle using
\begin{subequations}
\begin{align}
    \tan(\delta) &= \frac{2L sin(\alpha)}{l_{d}} \\
    l_d &= l_0 + \beta \lvert v \rvert 
\end{align}
\end{subequations}
Using the steering angle, we can calculate the angular velocity using

\begin{equation}
    \omega = \frac{v}{R} = \frac{v}{L} \tan(\delta) = \frac{2v sin(\alpha)}{l_{d}}
\end{equation}

Once we have the required velocity $v$ and angular velocity $\omega$, we can estimate the left and right wheel velocities using
\begin{subequations}
\begin{align}
    v_l &= v - \frac{\omega L}{2} \\
    v_r &= v + \frac{\omega L}{2} 
\end{align}
\end{subequations}

The standard pure pursuit algorithm only considers kinematic constraints and ignores dynamic forces on the tires and external disturbances which might cause instability especially in uneven terrains like agricultural fields. In order to address this issue we implement an adaptive Pure Pursuit algorithm similar to \citep{Hoffmann2007AutonomousAT} accounting for ground reactions causing side slips. An added gain term is added to the nominal input which acts as an active damper against the ground reaction forces causing slip. The modified steering control input is calculated as,

\begin{equation}
    \delta = \tan^{-1} \Big\{ \frac{2L sin(\alpha)}{l_{d}} \Big\} + k_{d,yaw}(r_{ref}-r)
\end{equation}
where $r_{ref}$ is the rate of change of yaw for the reference trajectory and $r$ is the change of yaw for the robot. The added gain of $k_{d,yaw}$ requires to be tuned based on the range of operating velocity of the robot.
\begin{figure}[!tbh]
    \centering
    \includegraphics[width = 0.45 \textwidth]{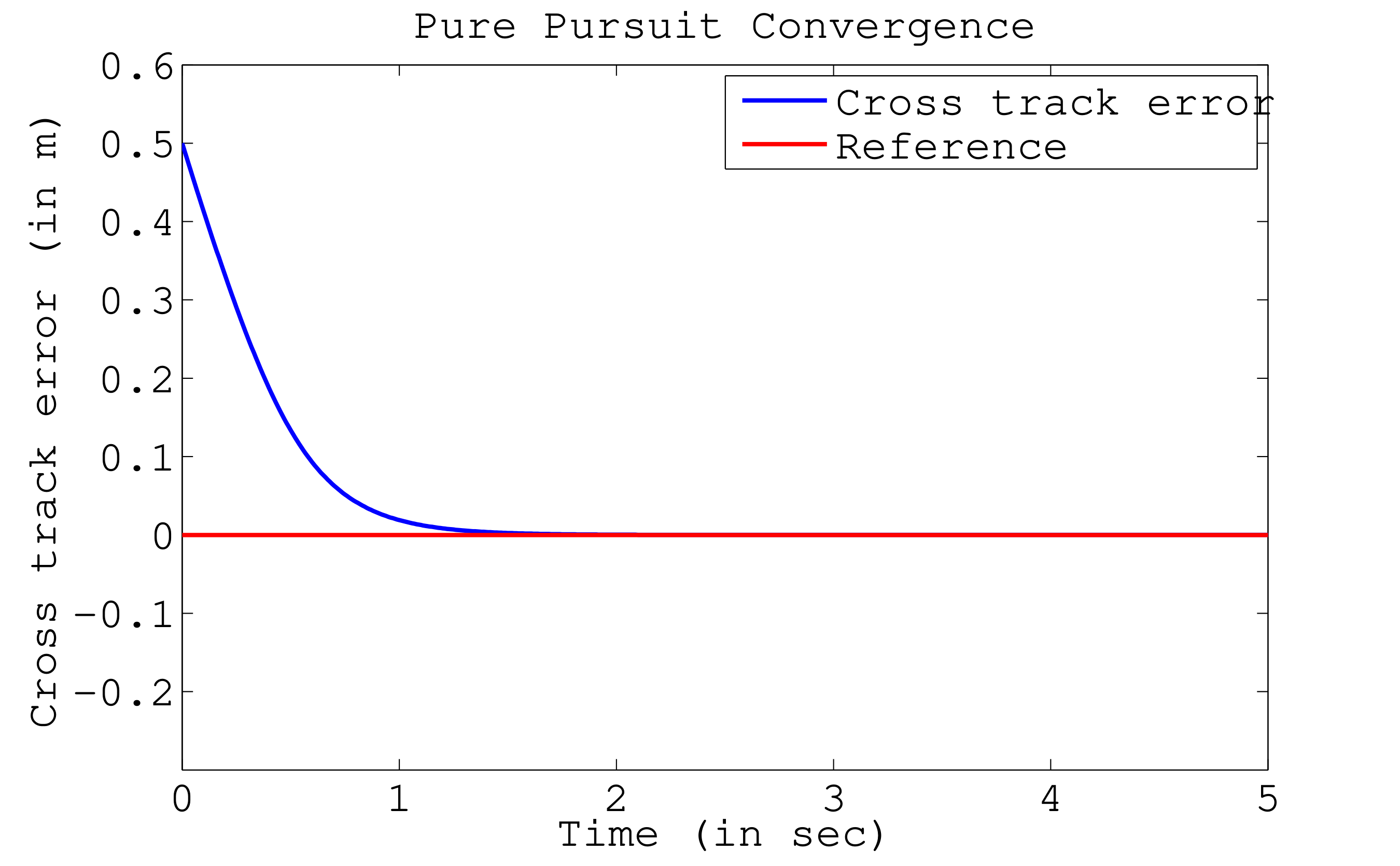}
    \caption{Convergence of Tracking Controller}
    \label{fig:pure_conv}
\end{figure}
The controller shows good performance in terrace farming tasks and converges to zero cross track error even in rough terrains. The Fig.\ref{fig:pure_conv} shows the convergence data of the controller implemented in a real-world testing environment.

\subsubsection{Farm Tool Control}
\paragraph{Pesticide Spraying and Watering:}
The robot possesses two different containers for pesticides and water which are connected to a DC motor-driven pump. On the basis of the agricultural strategy, the control system needs to provide the required actuation signal to the DC motor. Taking inspiration from the prior work in controlled pesticide spraying in \cite{Deshmukh2020DesignAD}, we employ solenoid valves to switch between the two tanks. The inlet of the DC motor pump is connected to the junction of the two pipes coming from the respective tanks through a solenoid valve, and the outlet of the pump is connected to the series nozzle system which can spray the pesticide or water based on the agriculture strategy. The solenoid valves have to be in a closed condition in order to allow flow only when they are actuated.

\paragraph{Ploughing, Sowing and Rolling:}
The distance between two seeds for sowing is controlled by a stepper. Let $n_t$, $r$, $\omega$ be the number of teeth, the inner radius of the seed feeding wheel, and the RPM of the stepper motor respectively. We denote the forward velocity of the robot at time $i\Delta t$ as $v_i$ (m/s), where $\Delta t$ is the sampling time of the control module. Given a desired seed separation of $d$ meters, and assuming one seed is being carried per tooth and a constant robot velocity for the next $\Delta t$ seconds, the seeder stepper motor RPM $\omega_i$ for this time step can be given by the Eq. \ref{eq_omega_i}. Therefore, the desired motor RPM is achieved by controlling the pulse speed $p$ [Hz] to the stepper motor to attain this desired angular velocity, given by Eq. \ref{eq_pulse_speed}.
\begin{align}
    \omega_i &= \frac{v_i}{n_t d} \label{eq_omega_i} \\
    p_i &= {\omega_i}{\alpha} \label{eq_pulse_speed}
\end{align}
Where $\alpha$ is the step angle of the stepper motor, with units rad/pulse.

\section{Conclusion}
\label{Conclusions}
This work describes Aarohi -- a fully automated solution to terrace farming, which is widely used for agriculture in hilly areas. The robot incorporates a novel stair climbing mechanism design, which facilitates climbing terrace steps of large heights. An optimisation-based design technique has been presented to determine optimal design parameters for the stair climbing mechanism based on the given use case. The agricultural module has been developed to perform farming activities like ploughing, sowing, rolling, watering and pesticide spraying efficiently and autonomously. A robust software stack has been developed, which uses a combination of 1D lidars, GPS and IMU sensors, wheel encoders and RGB camera images for robot localisation, task planning and execution. The proposed robot design is modular, and parameters like the number of wheel pairs, maximum stair climbing distance and maximum load-carrying capacity can be customised according to specific use cases or needs of the farmer. The prototype of this robot secured the First Position in the DIC’s Autonomous Terrace Farming event at the 8th Inter IIT Tech Meet, IIT Roorkee. We would like to conclude that from a commercial point of view, ‘Aarohi’ could bridge the current gap in extending automated farming to hilly regions, and prove to be a great solution to all terrace farming needs at an affordable price. 

\bmhead{Acknowledgments}
We would like to thank the MN Faruqui Innovations Center (MNFIC), IIT Kharagpur for the continuous support in developing and testing the prototype. We would also like to extend a special thanks to Mr. Utkarsh Agrawal, Mr. Himanshu Raj Khandelwal, Mr. Yash Agarwal, Mr. Aditya Rathore, Mr. Het Shah, Mr. Jaydeep Godbole and Mr. Utkarsh Sharma for their extensive efforts and contribution during the prototype development stage.

 
\newpage
\bibliography{sn-bibliography}


\begin{thebibliography}{71}
\ifx \bisbn   \undefined \def \bisbn  #1{ISBN #1}\fi
\ifx \binits  \undefined \def \binits#1{#1}\fi
\ifx \bauthor  \undefined \def \bauthor#1{#1}\fi
\ifx \batitle  \undefined \def \batitle#1{#1}\fi
\ifx \bjtitle  \undefined \def \bjtitle#1{#1}\fi
\ifx \bvolume  \undefined \def \bvolume#1{\textbf{#1}}\fi
\ifx \byear  \undefined \def \byear#1{#1}\fi
\ifx \bissue  \undefined \def \bissue#1{#1}\fi
\ifx \bfpage  \undefined \def \bfpage#1{#1}\fi
\ifx \blpage  \undefined \def \blpage #1{#1}\fi
\ifx \burl  \undefined \def \burl#1{\textsf{#1}}\fi
\ifx \doiurl  \undefined \def \doiurl#1{\url{https://doi.org/#1}}\fi
\ifx \betal  \undefined \def \betal{\textit{et al.}}\fi
\ifx \binstitute  \undefined \def \binstitute#1{#1}\fi
\ifx \binstitutionaled  \undefined \def \binstitutionaled#1{#1}\fi
\ifx \bctitle  \undefined \def \bctitle#1{#1}\fi
\ifx \beditor  \undefined \def \beditor#1{#1}\fi
\ifx \bpublisher  \undefined \def \bpublisher#1{#1}\fi
\ifx \bbtitle  \undefined \def \bbtitle#1{#1}\fi
\ifx \bedition  \undefined \def \bedition#1{#1}\fi
\ifx \bseriesno  \undefined \def \bseriesno#1{#1}\fi
\ifx \blocation  \undefined \def \blocation#1{#1}\fi
\ifx \bsertitle  \undefined \def \bsertitle#1{#1}\fi
\ifx \bsnm \undefined \def \bsnm#1{#1}\fi
\ifx \bsuffix \undefined \def \bsuffix#1{#1}\fi
\ifx \bparticle \undefined \def \bparticle#1{#1}\fi
\ifx \barticle \undefined \def \barticle#1{#1}\fi
\bibcommenthead
\ifx \bconfdate \undefined \def \bconfdate #1{#1}\fi
\ifx \botherref \undefined \def \botherref #1{#1}\fi
\ifx \url \undefined \def \url#1{\textsf{#1}}\fi
\ifx \bchapter \undefined \def \bchapter#1{#1}\fi
\ifx \bbook \undefined \def \bbook#1{#1}\fi
\ifx \bcomment \undefined \def \bcomment#1{#1}\fi
\ifx \oauthor \undefined \def \oauthor#1{#1}\fi
\ifx \citeauthoryear \undefined \def \citeauthoryear#1{#1}\fi
\ifx \endbibitem  \undefined \def \endbibitem {}\fi
\ifx \bconflocation  \undefined \def \bconflocation#1{#1}\fi
\ifx \arxivurl  \undefined \def \arxivurl#1{\textsf{#1}}\fi
\csname PreBibitemsHook\endcsname

\bibitem{GDP}
\begin{botherref}
Ministry of Statistics and Programme Implementation, Sector-wise GDP of India,
  Government of India.
\url{https://statisticstimes.com/economy/country/india-gdp-sectorwise.php}.
Accessed: 2021-07-10
(2021)
\end{botherref}
\endbibitem

\bibitem{engdawork2014long}
\begin{barticle}
\bauthor{\bsnm{Engdawork}, \binits{A.}},
\bauthor{\bsnm{Bork}, \binits{H.-R.}}:
\batitle{Long-term indigenous soil conservation technology in the chencha area,
  southern ethiopia: origin, characteristics, and sustainability}.
\bjtitle{Ambio}
\bvolume{43}(\bissue{7}),
\bfpage{932}--\blpage{942}
(\byear{2014})
\end{barticle}
\endbibitem

\bibitem{deng2021advantages}
\begin{botherref}
\oauthor{\bsnm{Deng}, \binits{C.}},
\oauthor{\bsnm{Zhang}, \binits{G.}},
\oauthor{\bsnm{Liu}, \binits{Y.}},
\oauthor{\bsnm{Nie}, \binits{X.}},
\oauthor{\bsnm{Li}, \binits{Z.}},
\oauthor{\bsnm{Liu}, \binits{J.}},
\oauthor{\bsnm{Zhu}, \binits{D.}}:
Advantages and disadvantages of terracing: A comprehensive review.
International Soil and Water Conservation Research
(2021)
\end{botherref}
\endbibitem

\bibitem{hamshere2014india}
\begin{botherref}
\oauthor{\bsnm{Hamshere}, \binits{P.}},
\oauthor{\bsnm{Sheng}, \binits{Y.}},
\oauthor{\bsnm{Moir}, \binits{B.}},
\oauthor{\bsnm{Gunning-Trant}, \binits{C.}},
\oauthor{\bsnm{Mobsby}, \binits{D.}}, et al.:
What india wants: analysis of india's food demand to 2050.
ABARES Research Report
(14.16)
(2014)
\end{botherref}
\endbibitem

\bibitem{mckinsey}
\begin{botherref}
\oauthor{\bsnm{McKinsey\phantom{ }and\phantom{ }Company}}:
Human + machine: A new era of automation in manufacturing.
Accessed: 2021-07-10
(2017)
\end{botherref}
\endbibitem

\bibitem{spugnoli2013environmental}
\begin{barticle}
\bauthor{\bsnm{Spugnoli}, \binits{P.}},
\bauthor{\bsnm{Dainelli}, \binits{R.}}:
\batitle{Environmental comparison of draught animal and tractor power}.
\bjtitle{Sustainability science}
\bvolume{8}(\bissue{1}),
\bfpage{61}--\blpage{72}
(\byear{2013})
\end{barticle}
\endbibitem

\bibitem{tiwari2004implications}
\begin{barticle}
\bauthor{\bsnm{Tiwari}, \binits{T.}},
\bauthor{\bsnm{Brook}, \binits{R.}},
\bauthor{\bsnm{Sinclair}, \binits{F.}}:
\batitle{Implications of hill farmers'agronomic practices in nepal for crop
  improvement in maize}.
\bjtitle{Experimental Agriculture}
\bvolume{40}(\bissue{4}),
\bfpage{397}--\blpage{417}
(\byear{2004})
\end{barticle}
\endbibitem

\bibitem{varisco1991future}
\begin{barticle}
\bauthor{\bsnm{Varisco}, \binits{D.M.}}:
\batitle{The future of terrace farming in yemen: a development dilemma}.
\bjtitle{Agriculture and Human Values}
\bvolume{8}(\bissue{1}),
\bfpage{166}--\blpage{172}
(\byear{1991})
\end{barticle}
\endbibitem

\bibitem{chapagain2017agronomic}
\begin{barticle}
\bauthor{\bsnm{Chapagain}, \binits{T.}},
\bauthor{\bsnm{Raizada}, \binits{M.N.}}:
\batitle{Agronomic challenges and opportunities for smallholder terrace
  agriculture in developing countries}.
\bjtitle{Frontiers in Plant Science}
\bvolume{8},
\bfpage{331}
(\byear{2017})
\end{barticle}
\endbibitem

\bibitem{10.1145/3387304.3387321}
\begin{bchapter}
\bauthor{\bsnm{Singhal}, \binits{A.}},
\bauthor{\bsnm{Mohta}, \binits{V.}},
\bauthor{\bsnm{Khandelwal}, \binits{Y.}},
\bauthor{\bsnm{Patnaik}, \binits{A.}},
\bauthor{\bsnm{Patel}, \binits{M.}},
\bauthor{\bsnm{Godbole}, \binits{J.}},
\bauthor{\bsnm{Priya}, \binits{S.}},
\bauthor{\bsnm{Dey}, \binits{S.}},
\bauthor{\bsnm{Shrivastava}, \binits{S.}},
\bauthor{\bsnm{Jhunjhunwala}, \binits{A.}},
\bauthor{\bsnm{Kowshik}, \binits{S.}},
\bauthor{\bsnm{Agrawal}, \binits{D.}},
\bauthor{\bsnm{Agarwal}, \binits{S.}},
\bauthor{\bsnm{Jha}, \binits{A.}},
\bauthor{\bsnm{Singh}, \binits{R.}},
\bauthor{\bsnm{Raj}, \binits{K.}},
\bauthor{\bsnm{Sahoo}, \binits{S.}},
\bauthor{\bsnm{Singh}, \binits{A.}},
\bauthor{\bsnm{Mallik}, \binits{R.}},
\bauthor{\bsnm{Lodhi}, \binits{V.}},
\bauthor{\bsnm{Chakravarty}, \binits{D.}}:
\bctitle{A prototype of an intelligent ground vehicle for constrained
  environment: Design and development}.
In: \bbtitle{Proceedings of the 2019 2nd International Conference on Control
  and Robot Technology}.
\bsertitle{ICCRT 2019},
pp. \bfpage{103}--\blpage{107}.
\bpublisher{Association for Computing Machinery},
\blocation{New York, NY, USA}
(\byear{2019})
\end{bchapter}
\endbibitem

\bibitem{9265493}
\begin{bchapter}
\bauthor{\bsnm{Krishnan}, \binits{S.V.}},
\bauthor{\bsnm{Singh}, \binits{V.}},
\bauthor{\bsnm{Shah}, \binits{P.}},
\bauthor{\bsnm{Yadav}, \binits{A.}},
\bauthor{\bsnm{Panampilly}, \binits{G.}},
\bauthor{\bsnm{Saha}, \binits{S.}},
\bauthor{\bsnm{Shukla}, \binits{H.D.}}:
\bctitle{Development of an rfid-based semi-autonomous robotic library
  management system}.
In: \bbtitle{2020 4th International Conference on Automation, Control and
  Robots (ICACR)},
pp. \bfpage{26}--\blpage{31}
(\byear{2020})
\end{bchapter}
\endbibitem

\bibitem{9476722}
\begin{bchapter}
\bauthor{\bsnm{Harinarayana}, \binits{T.}},
\bauthor{\bsnm{Krishnan}, \binits{S.V.}},
\bauthor{\bsnm{Hota}, \binits{S.}},
\bauthor{\bsnm{Kushwaha}, \binits{R.}}:
\bctitle{A lyapunov guidance vector field based continuous curvature path
  generation for waypoint following of uavs}.
In: \bbtitle{2021 International Conference on Unmanned Aircraft Systems
  (ICUAS)},
pp. \bfpage{498}--\blpage{506}
(\byear{2021})
\end{bchapter}
\endbibitem

\bibitem{singhal2019real}
\begin{bchapter}
\bauthor{\bsnm{Singhal}, \binits{A.}},
\bauthor{\bsnm{Mohta}, \binits{V.}},
\bauthor{\bsnm{Jha}, \binits{A.}},
\bauthor{\bsnm{Khandelwal}, \binits{Y.}},
\bauthor{\bsnm{Agrawal}, \binits{e.a.}}:
\bctitle{Real-time lane detection, fitting and navigation for unstructured
  environments}.
In: \bbtitle{2019 International Conference on Image and Video Processing, and
  Artificial Intelligence},
vol. \bseriesno{11321},
p. \bfpage{113210}
(\byear{2019})
\end{bchapter}
\endbibitem

\bibitem{Ladybird}
\begin{barticle}
\bauthor{\bsnm{Bender}, \binits{A.}},
\bauthor{\bsnm{Whelan}, \binits{B.}},
\bauthor{\bsnm{Sukkarieh}, \binits{S.}}:
\batitle{A high-resolution, multimodal data set for agricultural robotics: A
  ladybird's-eye view of brassica}.
\bjtitle{Journal of Field Robotics}
\bvolume{37}(\bissue{1}),
\bfpage{73}--\blpage{96}
(\byear{2020})
\end{barticle}
\endbibitem

\bibitem{xiong2020autonomous}
\begin{barticle}
\bauthor{\bsnm{Xiong}, \binits{Y.}},
\bauthor{\bsnm{Ge}, \binits{Y.}},
\bauthor{\bsnm{Grimstad}, \binits{L.}},
\bauthor{\bsnm{From}, \binits{P.J.}}:
\batitle{An autonomous strawberry-harvesting robot: Design, development,
  integration, and field evaluation}.
\bjtitle{Journal of Field Robotics}
\bvolume{37}(\bissue{2}),
\bfpage{202}--\blpage{224}
(\byear{2020})
\end{barticle}
\endbibitem

\bibitem{blender2016managing}
\begin{bchapter}
\bauthor{\bsnm{Blender}, \binits{T.}},
\bauthor{\bsnm{Buchner}, \binits{T.}},
\bauthor{\bsnm{Fernandez}, \binits{B.}},
\bauthor{\bsnm{Pichlmaier}, \binits{B.}},
\bauthor{\bsnm{Schlegel}, \binits{C.}}:
\bctitle{Managing a mobile agricultural robot swarm for a seeding task}.
In: \bbtitle{IECON 2016-42nd Annual Conference of the IEEE Industrial
  Electronics Society},
pp. \bfpage{6879}--\blpage{6886}
(\byear{2016}).
\bcomment{IEEE}
\end{bchapter}
\endbibitem

\bibitem{naik2016precision}
\begin{bchapter}
\bauthor{\bsnm{Naik}, \binits{N.S.}},
\bauthor{\bsnm{Shete}, \binits{V.V.}},
\bauthor{\bsnm{Danve}, \binits{S.R.}}:
\bctitle{Precision agriculture robot for seeding function}.
In: \bbtitle{2016 International Conference on Inventive Computation
  Technologies (ICICT)},
vol. \bseriesno{2},
pp. \bfpage{1}--\blpage{3}
(\byear{2016}).
\bcomment{IEEE}
\end{bchapter}
\endbibitem

\bibitem{arad2020development}
\begin{barticle}
\bauthor{\bsnm{Arad}, \binits{B.}},
\bauthor{\bsnm{Balendonck}, \binits{J.}},
\bauthor{\bsnm{Barth}, \binits{R.}},
\bauthor{\bsnm{Ben-Shahar}, \binits{O.}},
\bauthor{\bsnm{Edan}, \binits{Y.}},
\bauthor{\bsnm{Hellstr{\"o}m}, \binits{T.}},
\bauthor{\bsnm{Hemming}, \binits{J.}},
\bauthor{\bsnm{Kurtser}, \binits{P.}},
\bauthor{\bsnm{Ringdahl}, \binits{O.}},
\bauthor{\bsnm{Tielen}, \binits{T.}}, \betal:
\batitle{Development of a sweet pepper harvesting robot}.
\bjtitle{Journal of Field Robotics}
\bvolume{37}(\bissue{6}),
\bfpage{1027}--\blpage{1039}
(\byear{2020})
\end{barticle}
\endbibitem

\bibitem{silwal2017design}
\begin{barticle}
\bauthor{\bsnm{Silwal}, \binits{A.}},
\bauthor{\bsnm{Davidson}, \binits{J.R.}},
\bauthor{\bsnm{Karkee}, \binits{M.}},
\bauthor{\bsnm{Mo}, \binits{C.}},
\bauthor{\bsnm{Zhang}, \binits{Q.}},
\bauthor{\bsnm{Lewis}, \binits{K.}}:
\batitle{Design, integration, and field evaluation of a robotic apple
  harvester}.
\bjtitle{Journal of Field Robotics}
\bvolume{34}(\bissue{6}),
\bfpage{1140}--\blpage{1159}
(\byear{2017})
\end{barticle}
\endbibitem

\bibitem{berenstein2017human}
\begin{barticle}
\bauthor{\bsnm{Berenstein}, \binits{R.}},
\bauthor{\bsnm{Edan}, \binits{Y.}}:
\batitle{Human-robot collaborative site-specific sprayer}.
\bjtitle{Journal of Field Robotics}
\bvolume{34}(\bissue{8}),
\bfpage{1519}--\blpage{1530}
(\byear{2017})
\end{barticle}
\endbibitem

\bibitem{sowjanya2017multipurpose}
\begin{bchapter}
\bauthor{\bsnm{Sowjanya}, \binits{K.D.}},
\bauthor{\bsnm{Sindhu}, \binits{R.}},
\bauthor{\bsnm{Parijatham}, \binits{M.}},
\bauthor{\bsnm{Srikanth}, \binits{K.}},
\bauthor{\bsnm{Bhargav}, \binits{P.}}:
\bctitle{Multipurpose autonomous agricultural robot}.
In: \bbtitle{2017 International Conference of Electronics, Communication and
  Aerospace Technology (ICECA)},
vol. \bseriesno{2},
pp. \bfpage{696}--\blpage{699}
(\byear{2017}).
\bcomment{IEEE}
\end{bchapter}
\endbibitem

\bibitem{gollakota2011agribot}
\begin{bchapter}
\bauthor{\bsnm{Gollakota}, \binits{A.}},
\bauthor{\bsnm{Srinivas}, \binits{M.}}:
\bctitle{Agribot—a multipurpose agricultural robot}.
In: \bbtitle{2011 Annual IEEE India Conference},
pp. \bfpage{1}--\blpage{4}
(\byear{2011}).
\bcomment{IEEE}
\end{bchapter}
\endbibitem

\bibitem{kong2018autonomous}
\begin{bchapter}
\bauthor{\bsnm{Kong}, \binits{X.T.}},
\bauthor{\bsnm{Lu}, \binits{W.}},
\bauthor{\bsnm{Lu}, \binits{S.}}:
\bctitle{Autonomous drone-enabled terrace protection system}.
In: \bbtitle{2018 IEEE 15th International Conference on Networking, Sensing and
  Control (ICNSC)},
pp. \bfpage{1}--\blpage{6}
(\byear{2018}).
\bcomment{IEEE}
\end{bchapter}
\endbibitem

\bibitem{kulbacki2018survey}
\begin{bchapter}
\bauthor{\bsnm{Kulbacki}, \binits{M.}},
\bauthor{\bsnm{Segen}, \binits{J.}},
\bauthor{\bsnm{Knie{\'c}}, \binits{W.}},
\bauthor{\bsnm{Klempous}, \binits{R.}},
\bauthor{\bsnm{Kluwak}, \binits{K.}},
\bauthor{\bsnm{Nikodem}, \binits{J.}},
\bauthor{\bsnm{Kulbacka}, \binits{J.}},
\bauthor{\bsnm{Serester}, \binits{A.}}:
\bctitle{Survey of drones for agriculture automation from planting to harvest}.
In: \bbtitle{2018 IEEE 22nd International Conference on Intelligent Engineering
  Systems (INES)},
pp. \bfpage{000353}--\blpage{000358}
(\byear{2018}).
\bcomment{IEEE}
\end{bchapter}
\endbibitem

\bibitem{farmer_income}
\begin{bchapter}
\bauthor{\bsnm{Ranganathan}, \binits{T.}}:
\bctitle{Farmers’ income in india: Evidence from secondary data}.
(\byear{2015})
\end{bchapter}
\endbibitem

\bibitem{Siegwart1998DesignAI}
\begin{bchapter}
\bauthor{\bsnm{Siegwart}, \binits{R.}},
\bauthor{\bsnm{Lauria}, \binits{M.}},
\bauthor{\bsnm{M{\"a}usli}, \binits{P.}},
\bauthor{\bsnm{Winnendael}, \binits{M.V.}}:
\bctitle{Design and implementation of an innovative micro-rover}.
(\byear{1998})
\end{bchapter}
\endbibitem

\bibitem{Dalvand2006StairCS}
\begin{barticle}
\bauthor{\bsnm{Dalvand}, \binits{M.}},
\bauthor{\bsnm{Moghadam}, \binits{M.}}:
\batitle{Stair climber smart mobile robot (msrox)}.
\bjtitle{Autonomous Robots}
\bvolume{20},
\bfpage{3}--\blpage{14}
(\byear{2006})
\end{barticle}
\endbibitem

\bibitem{Lee2009ASR}
\begin{barticle}
\bauthor{\bsnm{Lee}, \binits{J.}},
\bauthor{\bsnm{Kim}, \binits{B.-S.}},
\bauthor{\bsnm{Song}, \binits{J.-B.}}:
\batitle{A small robot based on hybrid wheel-track mechanism}.
\bjtitle{Transactions of The Korean Society of Mechanical Engineers A}
\bvolume{33},
\bfpage{545}--\blpage{551}
(\byear{2009})
\end{barticle}
\endbibitem

\bibitem{Michaud2003AZIMUTAL}
\begin{barticle}
\bauthor{\bsnm{Michaud}, \binits{F.}},
\bauthor{\bsnm{L{\'e}tourneau}, \binits{D.}},
\bauthor{\bsnm{Arsenault}, \binits{M.}},
\bauthor{\bsnm{Bergeron}, \binits{Y.}},
\bauthor{\bsnm{Cadrin}, \binits{R.}},
\bauthor{\bsnm{Gagnon}, \binits{F.}},
\bauthor{\bsnm{Legault}, \binits{M.-A.}},
\bauthor{\bsnm{Millette}, \binits{M.}},
\bauthor{\bsnm{Par{\'e}}, \binits{J.}},
\bauthor{\bsnm{Tremblay}, \binits{M.-C.}},
\bauthor{\bsnm{Lepage}, \binits{P.}},
\bauthor{\bsnm{Morin}, \binits{Y.}},
\bauthor{\bsnm{Bisson}, \binits{J.}},
\bauthor{\bsnm{Caron}, \binits{S.}}:
\batitle{Azimut, a leg-track-wheel robot}.
\bjtitle{Proceedings 2003 IEEE/RSJ International Conference on Intelligent
  Robots and Systems (IROS 2003) (Cat. No.03CH37453)}
\bvolume{3},
\bfpage{2553}--\blpage{25583}
(\byear{2003})
\end{barticle}
\endbibitem

\bibitem{Kim2010WheelH}
\begin{bchapter}
\bauthor{\bsnm{Kim}, \binits{Y.-G.}},
\bauthor{\bsnm{Kim}, \binits{J.-W.}},
\bauthor{\bsnm{Kwak}, \binits{J.}},
\bauthor{\bsnm{Hong}, \binits{D.}},
\bauthor{\bsnm{Lee}, \binits{K.}},
\bauthor{\bsnm{An}, \binits{J.}}:
\bctitle{Wheel \&track hybrid mobile robot platform and mechanism for optimal
  navigation in urban terrain}.
(\byear{2010})
\end{bchapter}
\endbibitem

\bibitem{Yu2010ConfigurationAT}
\begin{botherref}
\oauthor{\bsnm{Yu}, \binits{S.}},
\oauthor{\bsnm{Wang}, \binits{T.}},
\oauthor{\bsnm{Li}, \binits{X.}},
\oauthor{\bsnm{Yao}, \binits{C.}},
\oauthor{\bsnm{Wang}, \binits{Z.}},
\oauthor{\bsnm{Zhi}, \binits{D.}}:
Configuration and tip-over stability analysis for stair-climbing of a new-style
  wheelchair robot.
2010 IEEE International Conference on Mechatronics and Automation,
1387--1392
(2010)
\end{botherref}
\endbibitem

\bibitem{Kim2012AutonomousTA}
\begin{barticle}
\bauthor{\bsnm{Kim}, \binits{Y.-G.}},
\bauthor{\bsnm{Kwak}, \binits{J.}},
\bauthor{\bsnm{Hong}, \binits{D.}},
\bauthor{\bsnm{Kim}, \binits{I.-H.}},
\bauthor{\bsnm{Shin}, \binits{D.}},
\bauthor{\bsnm{An}, \binits{J.}}:
\batitle{Autonomous terrain adaptation and user-friendly tele-operation of
  wheel-track hybrid mobile robot}.
\bjtitle{International Journal of Precision Engineering and Manufacturing}
\bvolume{13},
\bfpage{1781}--\blpage{1788}
(\byear{2012})
\end{barticle}
\endbibitem

\bibitem{1389593}
\begin{bchapter}
\bauthor{\bsnm{Gutmann}, \binits{J.-S.}},
\bauthor{\bsnm{Fukuchi}, \binits{M.}},
\bauthor{\bsnm{Fujita}, \binits{M.}}:
\bctitle{Stair climbing for humanoid robots using stereo vision}.
In: \bbtitle{2004 IEEE/RSJ International Conference on Intelligent Robots and
  Systems (IROS) (IEEE Cat. No.04CH37566)},
vol. \bseriesno{2},
pp. \bfpage{1407}--\blpage{14132}
(\byear{2004})
\end{bchapter}
\endbibitem

\bibitem{6094533}
\begin{bchapter}
\bauthor{\bsnm{Oßwald}, \binits{S.}},
\bauthor{\bsnm{Görög}, \binits{A.}},
\bauthor{\bsnm{Hornung}, \binits{A.}},
\bauthor{\bsnm{Bennewitz}, \binits{M.}}:
\bctitle{Autonomous climbing of spiral staircases with humanoids}.
In: \bbtitle{2011 IEEE/RSJ International Conference on Intelligent Robots and
  Systems},
pp. \bfpage{4844}--\blpage{4849}
(\byear{2011})
\end{bchapter}
\endbibitem

\bibitem{4399104}
\begin{bchapter}
\bauthor{\bsnm{Michel}, \binits{P.}},
\bauthor{\bsnm{Chestnutt}, \binits{J.}},
\bauthor{\bsnm{Kagami}, \binits{S.}},
\bauthor{\bsnm{Nishiwaki}, \binits{K.}},
\bauthor{\bsnm{Kuffner}, \binits{J.}},
\bauthor{\bsnm{Kanade}, \binits{T.}}:
\bctitle{Gpu-accelerated real-time 3d tracking for humanoid locomotion and
  stair climbing}.
In: \bbtitle{2007 IEEE/RSJ International Conference on Intelligent Robots and
  Systems},
pp. \bfpage{463}--\blpage{469}
(\byear{2007})
\end{bchapter}
\endbibitem

\bibitem{Saranl2001RHexAS}
\begin{barticle}
\bauthor{\bsnm{Saranlı}, \binits{U.}},
\bauthor{\bsnm{Buehler}, \binits{M.}},
\bauthor{\bsnm{Koditschek}, \binits{D.}}:
\batitle{Rhex: A simple and highly mobile hexapod robot}.
\bjtitle{The International Journal of Robotics Research}
\bvolume{20},
\bfpage{616}--\blpage{631}
(\byear{2001})
\end{barticle}
\endbibitem

\bibitem{Zhang2016ANC}
\begin{barticle}
\bauthor{\bsnm{Zhang}, \binits{L.}},
\bauthor{\bsnm{Yang}, \binits{Y.}},
\bauthor{\bsnm{Gu}, \binits{Y.}},
\bauthor{\bsnm{Xiaogang}, \binits{S.}},
\bauthor{\bsnm{Yao}, \binits{X.}},
\bauthor{\bsnm{Shuai}, \binits{L.}}:
\batitle{A new compact stair-cleaning robot}.
\bjtitle{Journal of Mechanisms and Robotics}
\bvolume{8},
\bfpage{045001}
(\byear{2016})
\end{barticle}
\endbibitem

\bibitem{scissor-lift}
\begin{botherref}
\oauthor{\bsnm{Saxena}, \binits{A.}}:
Deriving a generalized, actuator position-independent expression for the force
  output of a scissor lift
(2016)
\end{botherref}
\endbibitem

\bibitem{996017}
\begin{barticle}
\bauthor{\bsnm{Deb}, \binits{K.}},
\bauthor{\bsnm{Pratap}, \binits{A.}},
\bauthor{\bsnm{Agarwal}, \binits{S.}},
\bauthor{\bsnm{Meyarivan}, \binits{T.}}:
\batitle{A fast and elitist multiobjective genetic algorithm: Nsga-ii}.
\bjtitle{IEEE Transactions on Evolutionary Computation}
\bvolume{6}(\bissue{2}),
\bfpage{182}--\blpage{197}
(\byear{2002})
\end{barticle}
\endbibitem

\bibitem{mirjalili2016multi}
\begin{barticle}
\bauthor{\bsnm{Mirjalili}, \binits{S.}},
\bauthor{\bsnm{Saremi}, \binits{S.}},
\bauthor{\bsnm{Mirjalili}, \binits{S.M.}},
\bauthor{\bsnm{Coelho}, \binits{L.d.S.}}:
\batitle{Multi-objective grey wolf optimizer: a novel algorithm for
  multi-criterion optimization}.
\bjtitle{Expert Systems with Applications}
\bvolume{47},
\bfpage{106}--\blpage{119}
(\byear{2016})
\end{barticle}
\endbibitem

\bibitem{reyes2006multi}
\begin{barticle}
\bauthor{\bsnm{Reyes-Sierra}, \binits{M.}},
\bauthor{\bsnm{Coello}, \binits{C.C.}}, \betal:
\batitle{Multi-objective particle swarm optimizers: A survey of the
  state-of-the-art}.
\bjtitle{International journal of computational intelligence research}
\bvolume{2}(\bissue{3}),
\bfpage{287}--\blpage{308}
(\byear{2006})
\end{barticle}
\endbibitem

\bibitem{das2020multi}
\begin{barticle}
\bauthor{\bsnm{Das}, \binits{A.K.}},
\bauthor{\bsnm{Nikum}, \binits{A.K.}},
\bauthor{\bsnm{Krishnan}, \binits{S.V.}},
\bauthor{\bsnm{Pratihar}, \binits{D.K.}}:
\batitle{Multi-objective bonobo optimizer (mobo): an intelligent heuristic for
  multi-criteria optimization}.
\bjtitle{Knowledge and Information Systems}
\bvolume{62}(\bissue{11}),
\bfpage{4407}--\blpage{4444}
(\byear{2020})
\end{barticle}
\endbibitem

\bibitem{nagaraja2012plant}
\begin{barticle}
\bauthor{\bsnm{Nagaraja}, \binits{H.}},
\bauthor{\bsnm{Aswani}, \binits{R.}},
\bauthor{\bsnm{Malik}, \binits{M.}}:
\batitle{Plant watering autonomous mobile robot}.
\bjtitle{IAES International Journal of Robotics and Automation}
\bvolume{1}(\bissue{3}),
\bfpage{152}
(\byear{2012})
\end{barticle}
\endbibitem

\bibitem{chen2020design}
\begin{botherref}
\oauthor{\bsnm{Chen}, \binits{M.}},
\oauthor{\bsnm{Sun}, \binits{Y.}},
\oauthor{\bsnm{Cai}, \binits{X.}},
\oauthor{\bsnm{Liu}, \binits{B.}},
\oauthor{\bsnm{Ren}, \binits{T.}}:
Design and implementation of a novel precision irrigation robot based on an
  intelligent path planning algorithm.
arXiv preprint arXiv:2003.00676
(2020)
\end{botherref}
\endbibitem

\bibitem{bodunde2019architectural}
\begin{barticle}
\bauthor{\bsnm{Bodunde}, \binits{O.}},
\bauthor{\bsnm{Adie}, \binits{U.}},
\bauthor{\bsnm{Ikumapayi}, \binits{O.}},
\bauthor{\bsnm{Akinyoola}, \binits{J.}},
\bauthor{\bsnm{Aderoba}, \binits{A.}}:
\batitle{Architectural design and performance evaluation of a zigbee technology
  based adaptive sprinkler irrigation robot}.
\bjtitle{Computers and Electronics in Agriculture}
\bvolume{160},
\bfpage{168}--\blpage{178}
(\byear{2019})
\end{barticle}
\endbibitem

\bibitem{adeodu2019development}
\begin{barticle}
\bauthor{\bsnm{Adeodu}, \binits{A.}},
\bauthor{\bsnm{Bodunde}, \binits{O.}},
\bauthor{\bsnm{Daniyan}, \binits{I.}},
\bauthor{\bsnm{Omitola}, \binits{O.}},
\bauthor{\bsnm{Akinyoola}, \binits{J.}},
\bauthor{\bsnm{Adie}, \binits{U.}}:
\batitle{Development of an autonomous mobile plant irrigation robot for semi
  structured environment}.
\bjtitle{Procedia Manufacturing}
\bvolume{35},
\bfpage{9}--\blpage{15}
(\byear{2019})
\end{barticle}
\endbibitem

\bibitem{rafi2016design}
\begin{bchapter}
\bauthor{\bsnm{Rafi}, \binits{R.H.}},
\bauthor{\bsnm{Das}, \binits{S.}},
\bauthor{\bsnm{Ahmed}, \binits{N.}},
\bauthor{\bsnm{Hossain}, \binits{I.}},
\bauthor{\bsnm{Reza}, \binits{S.T.}}:
\bctitle{Design \& implementation of a line following robot for irrigation
  based application}.
In: \bbtitle{2016 19th International Conference on Computer and Information
  Technology (ICCIT)},
pp. \bfpage{480}--\blpage{483}
(\byear{2016}).
\bcomment{IEEE}
\end{bchapter}
\endbibitem

\bibitem{yallappa2017development}
\begin{bchapter}
\bauthor{\bsnm{Yallappa}, \binits{D.}},
\bauthor{\bsnm{Veerangouda}, \binits{M.}},
\bauthor{\bsnm{Maski}, \binits{D.}},
\bauthor{\bsnm{Palled}, \binits{V.}},
\bauthor{\bsnm{Bheemanna}, \binits{M.}}:
\bctitle{Development and evaluation of drone mounted sprayer for pesticide
  applications to crops}.
In: \bbtitle{2017 IEEE Global Humanitarian Technology Conference (GHTC)},
pp. \bfpage{1}--\blpage{7}
(\byear{2017}).
\bcomment{IEEE}
\end{bchapter}
\endbibitem

\bibitem{xue2016develop}
\begin{barticle}
\bauthor{\bsnm{Xue}, \binits{X.}},
\bauthor{\bsnm{Lan}, \binits{Y.}},
\bauthor{\bsnm{Sun}, \binits{Z.}},
\bauthor{\bsnm{Chang}, \binits{C.}},
\bauthor{\bsnm{Hoffmann}, \binits{W.C.}}:
\batitle{Develop an unmanned aerial vehicle based automatic aerial spraying
  system}.
\bjtitle{Computers and electronics in agriculture}
\bvolume{128},
\bfpage{58}--\blpage{66}
(\byear{2016})
\end{barticle}
\endbibitem

\bibitem{huang2015development}
\begin{barticle}
\bauthor{\bsnm{Huang}, \binits{Y.}},
\bauthor{\bsnm{Hoffman}, \binits{W.C.}},
\bauthor{\bsnm{Lan}, \binits{Y.}},
\bauthor{\bsnm{Fritz}, \binits{B.K.}},
\bauthor{\bsnm{Thomson}, \binits{S.J.}}:
\batitle{Development of a low-volume sprayer for an unmanned helicopter}.
\bjtitle{Journal of Agricultural Science}
\bvolume{7}(\bissue{1}),
\bfpage{148}
(\byear{2015})
\end{barticle}
\endbibitem

\bibitem{geng2012assessment}
\begin{barticle}
\bauthor{\bsnm{Geng}, \binits{C.}},
\bauthor{\bsnm{Zhang}, \binits{K.}},
\bauthor{\bsnm{Zhang}, \binits{E.}},
\bauthor{\bsnm{Zhang}, \binits{J.}},
\bauthor{\bsnm{Li}, \binits{W.}}:
\batitle{Assessment on spraying effect of intelligent spraying robot by
  experiment}.
\bjtitle{Transactions of the Chinese Society of Agricultural Engineering}
\bvolume{28}(\bissue{1}),
\bfpage{114}--\blpage{117}
(\byear{2012})
\end{barticle}
\endbibitem

\bibitem{sammons2005autonomous}
\begin{bchapter}
\bauthor{\bsnm{Sammons}, \binits{P.J.}},
\bauthor{\bsnm{Furukawa}, \binits{T.}},
\bauthor{\bsnm{Bulgin}, \binits{A.}}:
\bctitle{Autonomous pesticide spraying robot for use in a greenhouse}.
In: \bbtitle{Australian Conference on Robotics and Automation},
vol. \bseriesno{1}
(\byear{2005})
\end{bchapter}
\endbibitem

\bibitem{oberti2016selective}
\begin{barticle}
\bauthor{\bsnm{Oberti}, \binits{R.}},
\bauthor{\bsnm{Marchi}, \binits{M.}},
\bauthor{\bsnm{Tirelli}, \binits{P.}},
\bauthor{\bsnm{Calcante}, \binits{A.}},
\bauthor{\bsnm{Iriti}, \binits{M.}},
\bauthor{\bsnm{Tona}, \binits{E.}},
\bauthor{\bsnm{Ho{\v{c}}evar}, \binits{M.}},
\bauthor{\bsnm{Baur}, \binits{J.}},
\bauthor{\bsnm{Pfaff}, \binits{J.}},
\bauthor{\bsnm{Sch{\"u}tz}, \binits{C.}}, \betal:
\batitle{Selective spraying of grapevines for disease control using a modular
  agricultural robot}.
\bjtitle{Biosystems engineering}
\bvolume{146},
\bfpage{203}--\blpage{215}
(\byear{2016})
\end{barticle}
\endbibitem

\bibitem{bechar2016agricultural}
\begin{barticle}
\bauthor{\bsnm{Bechar}, \binits{A.}},
\bauthor{\bsnm{Vigneault}, \binits{C.}}:
\batitle{Agricultural robots for field operations: Concepts and components}.
\bjtitle{Biosystems Engineering}
\bvolume{149},
\bfpage{94}--\blpage{111}
(\byear{2016})
\end{barticle}
\endbibitem

\bibitem{bhattacharyya2020design}
\begin{botherref}
\oauthor{\bsnm{Bhattacharyya}, \binits{N.}}:
Design and development of intelligent pesticide spraying system for
  agricultural robot.
In: Hybrid Intelligent Systems: 20th International Conference on Hybrid
  Intelligent Systems (HIS 2020), December 14--16, 2020,
p. 157.
Springer Nature
\end{botherref}
\endbibitem

\bibitem{smith1976farm}
\begin{botherref}
\oauthor{\bsnm{Smith}, \binits{A.E.}},
\oauthor{\bsnm{Wilkes}, \binits{L.H.}}:
Farm machinery and equipment, 6th ed. New York: McGraw-Hill Book Company.
Inc
(1976)
\end{botherref}
\endbibitem

\bibitem{CHANDRAMOULI201823702}
\begin{barticle}
\bauthor{\bsnm{{Chandra Mouli}}, \binits{K.}},
\bauthor{\bsnm{Arunkumar}, \binits{S.}},
\bauthor{\bsnm{Satwik}, \binits{B.}},
\bauthor{\bsnm{Ram}, \binits{S.B.}},
\bauthor{\bsnm{{Rushi Tej}}, \binits{J.}},
\bauthor{\bsnm{Chaitanya}, \binits{A.S.}}:
\batitle{Design of reversible plough attachment}.
\bjtitle{Materials Today: Proceedings}
\bvolume{5}(\bissue{11, Part 3}),
\bfpage{23702}--\blpage{23709}
(\byear{2018})
\end{barticle}
\endbibitem

\bibitem{kepner1990principle}
\begin{botherref}
\oauthor{\bsnm{Kepner}, \binits{R.}},
\oauthor{\bsnm{Bainer}, \binits{R.}},
\oauthor{\bsnm{Berger}, \binits{E.}}:
Principle of farm machinery (3 rd eds) avi. pub.
Inc. West port Connecticut, USA. Pp122-163
(1990)
\end{botherref}
\endbibitem

\bibitem{osman2011effects}
\begin{barticle}
\bauthor{\bsnm{Osman}, \binits{A.N.}},
\bauthor{\bsnm{Xia}, \binits{L.}},
\bauthor{\bsnm{Dongxing}, \binits{Z.}}:
\batitle{Effects of tilt angle of disk plough on some soil physical properties,
  work rate and wheel slippage under light clay soil}.
\bjtitle{International Journal of Agricultural and Biological Engineering}
\bvolume{4}(\bissue{2}),
\bfpage{29}--\blpage{35}
(\byear{2011})
\end{barticle}
\endbibitem

\bibitem{bukhari1992effect}
\begin{barticle}
\bauthor{\bsnm{Bukhari}, \binits{S.}},
\bauthor{\bsnm{Mari}, \binits{G.}},
\bauthor{\bsnm{Zafaruhah}, \binits{M.}},
\bauthor{\bsnm{Baloch}, \binits{J.}},
\bauthor{\bsnm{Panhwar}, \binits{M.}}:
\batitle{Effect of disk and tilt angle on field capacity and power requirements
  of mounted plow}.
\bjtitle{Agricultural Mechanization in Asia, Africa, and Latin America}
\bvolume{23}(\bissue{2}),
\bfpage{9}--\blpage{13}
(\byear{1992})
\end{barticle}
\endbibitem

\bibitem{article_el_naim}
\begin{botherref}
\oauthor{\bsnm{El~Naim}, \binits{A.}},
\oauthor{\bsnm{Abdalla}, \binits{O.}},
\oauthor{\bsnm{Mohamed}, \binits{E.}},
\oauthor{\bsnm{Shiekh}, \binits{M.}},
\oauthor{\bsnm{Zaied}, \binits{M.}}:
Effect of disc and tilt angles of disc plough on tractor performance under clay
  soil.
Current Research in Agricultural Sciences, 2014, 1(3): 83-94
(2014)
\end{botherref}
\endbibitem

\bibitem{singh2005optimisation}
\begin{barticle}
\bauthor{\bsnm{Singh}, \binits{R.}},
\bauthor{\bsnm{Singh}, \binits{G.}},
\bauthor{\bsnm{Saraswat}, \binits{D.}}:
\batitle{Optimisation of design and operational parameters of a pneumatic seed
  metering device for planting cottonseeds}.
\bjtitle{Biosystems engineering}
\bvolume{92}(\bissue{4}),
\bfpage{429}--\blpage{438}
(\byear{2005})
\end{barticle}
\endbibitem

\bibitem{yasir2012design}
\begin{barticle}
\bauthor{\bsnm{Yasir}, \binits{S.H.}},
\bauthor{\bsnm{Liao}, \binits{Q.}},
\bauthor{\bsnm{Yu}, \binits{J.}},
\bauthor{\bsnm{He}, \binits{D.}}:
\batitle{Design and test of a pneumatic precision metering device for wheat}.
\bjtitle{Agricultural Engineering International: CIGR Journal}
\bvolume{14}(\bissue{1}),
\bfpage{16}--\blpage{25}
(\byear{2012})
\end{barticle}
\endbibitem

\bibitem{minfeng2018optimal}
\begin{barticle}
\bauthor{\bsnm{Minfeng}, \binits{J.}},
\bauthor{\bsnm{Yongqian}, \binits{D.}},
\bauthor{\bsnm{Hongfeng}, \binits{Y.}},
\bauthor{\bsnm{Haitao}, \binits{L.}},
\bauthor{\bsnm{Yizhuo}, \binits{J.}},
\bauthor{\bsnm{Xiuqing}, \binits{F.}}:
\batitle{Optimal structure design and performance tests of seed metering device
  with fluted rollers for precision wheat seeding machine}.
\bjtitle{IFAC-PapersOnLine}
\bvolume{51}(\bissue{17}),
\bfpage{509}--\blpage{514}
(\byear{2018})
\end{barticle}
\endbibitem

\bibitem{yu2021design}
\begin{bchapter}
\bauthor{\bsnm{Yu}, \binits{X.}},
\bauthor{\bsnm{Zhang}, \binits{B.}},
\bauthor{\bsnm{You}, \binits{J.}}:
\bctitle{Design and analysis of film-covering direct seeding machine in paddy
  field}.
In: \bbtitle{Journal of Physics: Conference Series},
vol. \bseriesno{1744},
p. \bfpage{022126}
(\byear{2021}).
\bcomment{IOP Publishing}
\end{bchapter}
\endbibitem

\bibitem{article_ground_wheel}
\begin{barticle}
\bauthor{\bsnm{Singh}, \binits{T.}},
\bauthor{\bsnm{Mane}, \binits{D.M.}}:
\batitle{Development and laboratory performance of an electronically controlled
  metering mechanism for okra seed}.
\bjtitle{AMA, Agricultural Mechanization in Asia, Africa and Latin America}
\bvolume{42},
\bfpage{63}--\blpage{69}
(\byear{2011})
\end{barticle}
\endbibitem

\bibitem{jianbo2014design}
\begin{barticle}
\bauthor{\bsnm{Jianbo}, \binits{Z.}},
\bauthor{\bsnm{Junfang}, \binits{X.}},
\bauthor{\bsnm{Yong}, \binits{Z.}},
\bauthor{\bsnm{Shun}, \binits{Z.}}:
\batitle{Design and experimental study of the control system for precision
  seed-metering device}.
\bjtitle{International Journal of Agricultural and Biological Engineering}
\bvolume{7}(\bissue{3}),
\bfpage{13}--\blpage{18}
(\byear{2014})
\end{barticle}
\endbibitem

\bibitem{seo2014vision}
\begin{bchapter}
\bauthor{\bsnm{Seo}, \binits{Y.-W.}},
\bauthor{\bsnm{Rajkumar}, \binits{R.}}:
\bctitle{A vision system for detecting and tracking of stop-lines}.
In: \bbtitle{17th International IEEE Conference on Intelligent Transportation
  Systems (ITSC)},
pp. \bfpage{1970}--\blpage{1975}
(\byear{2014}).
\bcomment{IEEE}
\end{bchapter}
\endbibitem

\bibitem{Alonso2013SelftuningPC}
\begin{botherref}
\oauthor{\bsnm{Alonso}, \binits{L.}},
\oauthor{\bsnm{P{\'e}rez-Oria}, \binits{J.}},
\oauthor{\bsnm{Al-Hadithi}, \binits{B.M.}},
\oauthor{\bsnm{Jim{\'e}nez}, \binits{A.}}:
Self-tuning pid controller for autonomous car tracking in urban traffic.
2013 17th International Conference on System Theory, Control and Computing
  (ICSTCC),
15--20
(2013)
\end{botherref}
\endbibitem

\bibitem{Hoffmann2007AutonomousAT}
\begin{botherref}
\oauthor{\bsnm{Hoffmann}, \binits{G.M.}},
\oauthor{\bsnm{Tomlin}, \binits{C.J.}},
\oauthor{\bsnm{Montemerlo}, \binits{M.D.}},
\oauthor{\bsnm{Thrun}, \binits{S.}}:
Autonomous automobile trajectory tracking for off-road driving: Controller
  design, experimental validation and racing.
2007 American Control Conference,
2296--2301
(2007)
\end{botherref}
\endbibitem

\bibitem{Deshmukh2020DesignAD}
\begin{bchapter}
\bauthor{\bsnm{Deshmukh}, \binits{D.}},
\bauthor{\bsnm{Pratihar}, \binits{D.K.}},
\bauthor{\bsnm{Deb}, \binits{A.K.}},
\bauthor{\bsnm{Ray}, \binits{H.}},
\bauthor{\bsnm{Bhattacharyya}, \binits{N.}}:
\bctitle{Design and development of intelligent pesticide spraying system for
  agricultural robot}.
In: \bbtitle{HIS}
(\byear{2020})
\end{bchapter}
\endbibitem

\end{thebibliography}

\end{document}